\pdfoutput=1
\documentclass[11pt]{article}

\usepackage[preprint]{acl}

\usepackage{times}
\usepackage{latexsym}
\usepackage{amsmath}
\usepackage{makecell}
\usepackage{enumitem}
\usepackage{amsmath}
\definecolor{lightgray}{gray}{0.9}
\definecolor{lightblue}{RGB}{0,191,255}

\usepackage[T1]{fontenc}
\usepackage[utf8]{inputenc}

\usepackage{microtype}

\usepackage{inconsolata}

\usepackage{graphicx}
\usepackage{amssymb}
\usepackage{amsfonts,cases,pifont}
\usepackage{booktabs,algorithm}
\usepackage{pgfplots}
\pgfplotsset{compat=1.18}
\usepackage{listings}
\usepackage{multirow}

\lstdefinestyle{custompython}{
    language=Python,
    basicstyle=\ttfamily\footnotesize,
    keywordstyle=\color{blue},
    commentstyle=\color{green!50!black},
    stringstyle=\color{red},
    numbers=left,
    numberstyle=\tiny\color{gray},
    stepnumber=1,
    numbersep=5pt,
    backgroundcolor=\color{gray!10},
    showspaces=false,
    showstringspaces=false,
    showtabs=false,
    tabsize=2
}
\title{Optimizing Singular Spectrum  for Large Language Model Compression}   % Task-aware

\author{
    \textbf{Dengjie Li\textsuperscript{*}, Tiancheng Shen\textsuperscript{*}, Yao Zhou, Bernard Ghanem, Yibo Yang, Ming-Hsuan Yang}
}

\begin{document}
\maketitle

\renewcommand{\thefootnote}{\fnsymbol{footnote}}
\footnotetext[1]{Equal contribution.}
\renewcommand{\thefootnote}{\arabic{footnote}}

\begin{abstract}
Large language models~(LLMs) have demonstrated remarkable capabilities, yet prohibitive parameter complexity often hinders their deployment. %computational and memory requirements. 
Existing singular value decomposition~(SVD) based compression methods simply deem singular values as importance scores of decomposed components. 
However, this importance ordered by singular values does not necessarily correlate with the performance of a downstream task. 
In this work, we introduce SoCo~(\textbf{S}ingular spectrum \textbf{o}ptimization for large language model \textbf{Co}mpression), a novel compression framework that learns to rescale the decomposed components of SVD in a data-driven manner. 
Concretely, we employ a learnable diagonal matrix to assign importance scores for singular spectrum and develop a three-stage training process that progressively refines these scores—from initial coarse compression to fine-grained sparsification—thereby striking an effective balance between aggressive model compression and performance preservation. 
Thanks to the learnable singular spectrum, SoCo adaptively prunes components according to the sparsified importance scores, rather than relying on the fixed order of singular values.
More importantly, the remaining components with amplified importance scores can compensate for the loss of the pruned ones.
%, enabling more nuanced and task-aware parameter pruning. 
Experimental evaluations across multiple LLMs and benchmarks demonstrate that SoCo surpasses the state-of-the-art methods in model compression. 
% Notably, compared with SVD-LLM, SoCo gains improvements of up to 37.6\%, 81.8\%, 94.8\%, and 99.2\% in performance at 20\%, 40\%, 60\%, 80\% compression rates on the C4 dataset.

%, while delivering performance on par with the state-of-the-art SVD-LLM~\cite{wang2024svd}, our proposed method SoCo can further compress the model by a factor of more than 3.
\end{abstract}

\begin{figure}[ht]
\begin{center}
\centerline{\includegraphics[width=1.0\columnwidth]{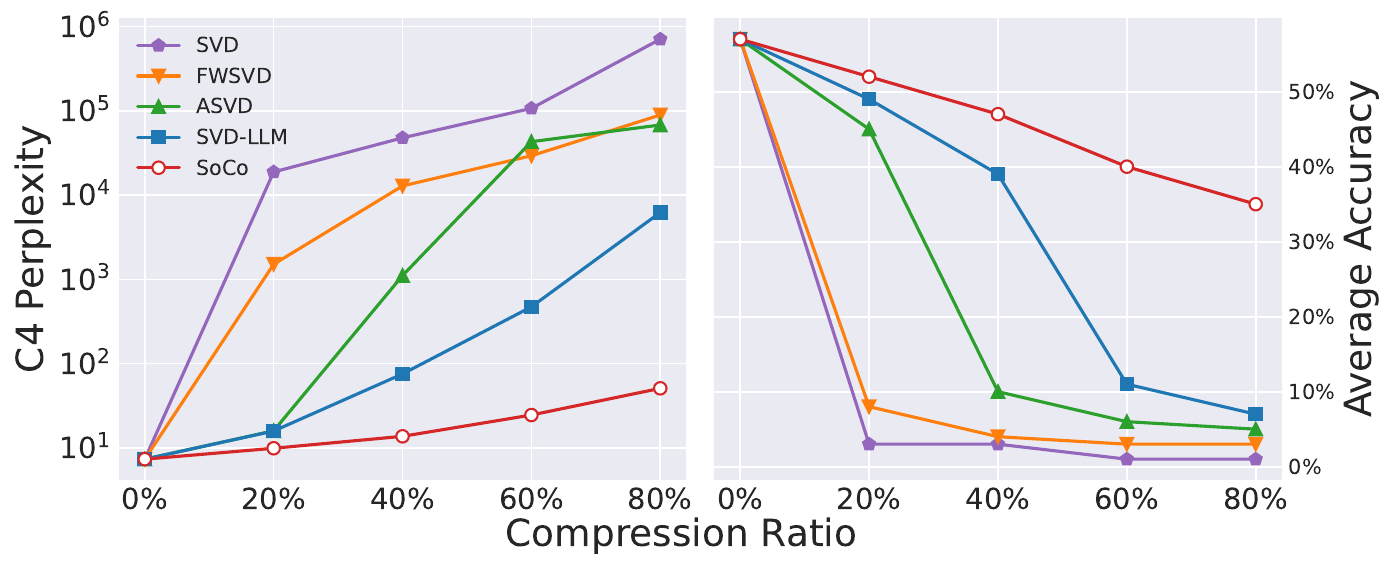}}
\end{center}
\vspace{-9mm}
\caption{SoCo consistently outperforms existing methods across a range of compression ratios, yielding lower perplexity on C4~(left) and higher average classification accuracy on LM-Evaluation-Harness~(right).}
\label{fig:intro}
\end{figure}

\section{Introduction}

Recent advances in large language models~(LLMs) have led to remarkable breakthroughs in natural language processing~(NLP) tasks.~\cite{waswani2017attention, devlin2018bert, brown2020Language, kaplan2020scaling, touvron2023llama}
However, the vast number of parameters in these deep transformer architectures results in prohibitive storage and computational requirements and poses significant challenges for deployment and inference efficiency~\cite{wan2023efficient, zhu2023survey}. 
Consequently, compression of LLMs has emerged as a vital problem~\cite{dettmers2022gpt3, ma2023llm, zhong2024revisiting, Hsu2022FWSVD}.

A classical approach to model compression performs singular value decomposition~(SVD)~\cite{stewart1993early, wall2003singular} for weight matrices and only preserves the principal components with the largest several singular values, which consist of low-rank matrices and largely reduce the number of parameters. 
These SVD-based methods assume that the decomposed components with smaller singular values are less important for model performance.
However, the order of singular values only reflects their contribution to low-rank approximation but does not necessarily correlate with the performance of a downstream task.
To address this, recent works, such as ASVD~\cite{yuan2023asvd} and SVD-LLM~\cite{wang2024svd}, introduce activation awareness with calibration data into the decomposition process to mitigate such mismatch. 
Nonetheless, these methods still perform truncation by the order of singular values without re-evaluating the intrinsic importance of their corresponding components.
As a result, the model performance will be rapidly unacceptable when the compression ratio increases. For example, as shown in Fig.~\ref{fig:intro}, when compression ratio is larger than 40\%, all the compared methods including original SVD~\cite{wall2003singular}, FWSVD~\cite{Hsu2022FWSVD}, ASVD~\cite{yuan2023asvd}, and SVD-LLM~\cite{wang2024svd} lead to catastrophic perplexity on C4~\cite{Raffel2019C4}. Besides, most existing methods cannot compensate for the loss caused by the pruned components~\cite{Hsu2022FWSVD, yuan2023asvd}. 

In this paper, we propose \textbf{SoCo}~(\textbf{S}ingular spectrum \textbf{o}ptimization for large language model \textbf{Co}mpression), a novel framework that learns to rescale the singular spectrum in a data-driven manner.
Concretely, SoCo introduces a learnable diagonal matrix in which each element contributes an importance score to re-evaluate the corresponding component. 
%in terms of both model performance and parameter complexity. 
We develop a three-stage training process to deal with the competing objectives of aggressive parameter reduction and performance preservation.  
In Stage 1, the model undergoes rapid, coarse compression to establish initial importance scores and reach the target compression ratio. 
Stage 2 uses an alternating optimization strategy to refine these scores, allowing the borderline components to be thoroughly evaluated through controlled oscillation around the target ratio.
Stage 3 enforces sparsity in the scores, clearly distinguishing essential components from those that can be pruned. 

The whole training process is efficient since we keep all the model parameters frozen. 
After training, we prune the components according to the sparsified importance scores, rather than simply truncating by the order of singular values as adopted in prior studies. 
This adaptive, data-aware strategy overcomes the task-agnostic evaluation of singular values and allows for a nuanced selection of the decomposed components. 
More importantly, the preserved components whose importance scores are amplified can compensate for the loss of pruning, further improving the compression-performance trade-off.
The main contributions of this work are:
\begin{itemize}[leftmargin=1em, itemsep=0.2ex, topsep=0.5ex, parsep=0ex]
\item We propose SoCo, a learning-based compression framework that optimizes the singular value spectrum through trainable importance scores to overcome the mismatch between the order of singular values from SVD and the true contribution to downstream task performance. 
\item We design a three-stage training strategy that gradually refines the importance scores through initial coarse compression, midterm adjustment, and final sparsification, resulting in a superior compression-performance trade-off. 
The preserved components with amplified importance scores can further compensate for the performance loss caused by pruning.  
\item 
Extensive evaluations with six popular LLMs~(from 7B to 30B parameters) across various benchmarks in Tab.~\ref{tab:svd_methods} demonstrate that SoCo achieves better compression performance than the state-of-the-art methods. Notably, compared with SVD-LLM~\cite{wang2024svd}, SoCo gains improvements of up to 37.6\%, 81.8\%, 94.8\%, and 99.2\% in performance at 20\%, 40\%, 60\%, 80\% compression rates on the C4 dataset~\cite{Raffel2019C4}.
\end{itemize}

\section{Related Work}

%\subsection{Large Language Model Compression}
\noindent \textbf{Large Language Model Compression.}
LLM compression methods aim to make LLMs more efficient for deployment~\cite{wan2023efficient, zhu2023survey}. Common techniques include quantization, pruning, knowledge distillation, and low-rank decomposition. Quantization reduces the precision of model weights and activations, thereby decreasing memory usage and computational load.~\cite{dettmers2022gpt3, kimsqueezellm, shen2024efficient,Lin2023AWQAW}.
Pruning involves removing less significant weights or neurons from the model to create a sparser architecture. However, pruning often necessitates retraining to recover potential performance degradation~\cite{ma2023llm, ashkboos2024slicegpt, zhong2024blockpruner, hsieh2023distilling}. Knowledge distillation transfers knowledge from a large ``teacher'' model to a smaller ``student'' model by training the latter to replicate the former's outputs. This approach enables the student model to achieve performance comparable to the teacher model while being more efficient~\cite{zhong2024revisiting, muralidharan2024compact, hsieh2023distilling}. Low-rank decomposition compresses weight matrices by factorizing them into smaller, computationally efficient components~\cite{Hsu2022FWSVD, yuan2023asvd, wang2024svd}.

%\subsection{Low-rank Decomposition Compression}
\vspace{1mm}
\noindent \textbf{Low-rank Decomposition Compression.}
A significant portion of existing low-rank compression techniques~\cite{li2023losparse, noach2020compressing, chen2023ternary, Hsu2022FWSVD, yuan2023asvd, wang2024svd} relies on Singular Value Decomposition~(SVD)~\cite{stewart1993early, wall2003singular} to factorize and compress model parameters.
% , thereby reducing model size while preserving performance. 
Although original SVD~\cite{stewart1993early, wall2003singular} decomposes weight matrices into lower-rank components for reconstruction, its objective may lead to information loss if not tailored for downstream tasks. Improved SVD-based methods address these limitations. For example, FWSVD~\cite{Hsu2022FWSVD} leverages Fisher information to preserve critical components, ASVD~\cite{yuan2023asvd} scales singular values based on activation sensitivity, and SVD-LLM~\cite{wang2024svd} incorporates data whitening and layer-wise updates to compensate for accuracy loss at high compression ratios.

However, these approaches depend on a fixed descending order of singular values from SVD, allowing only truncation adjustments without re-evaluating the true significance of each decomposed component. In contrast, our method introduces a learnable mechanism that re-evaluates component importance end-to-end, enabling a more adaptive and effective compression strategy.

\begin{figure*}[ht]
\centering
\includegraphics[width=\textwidth]{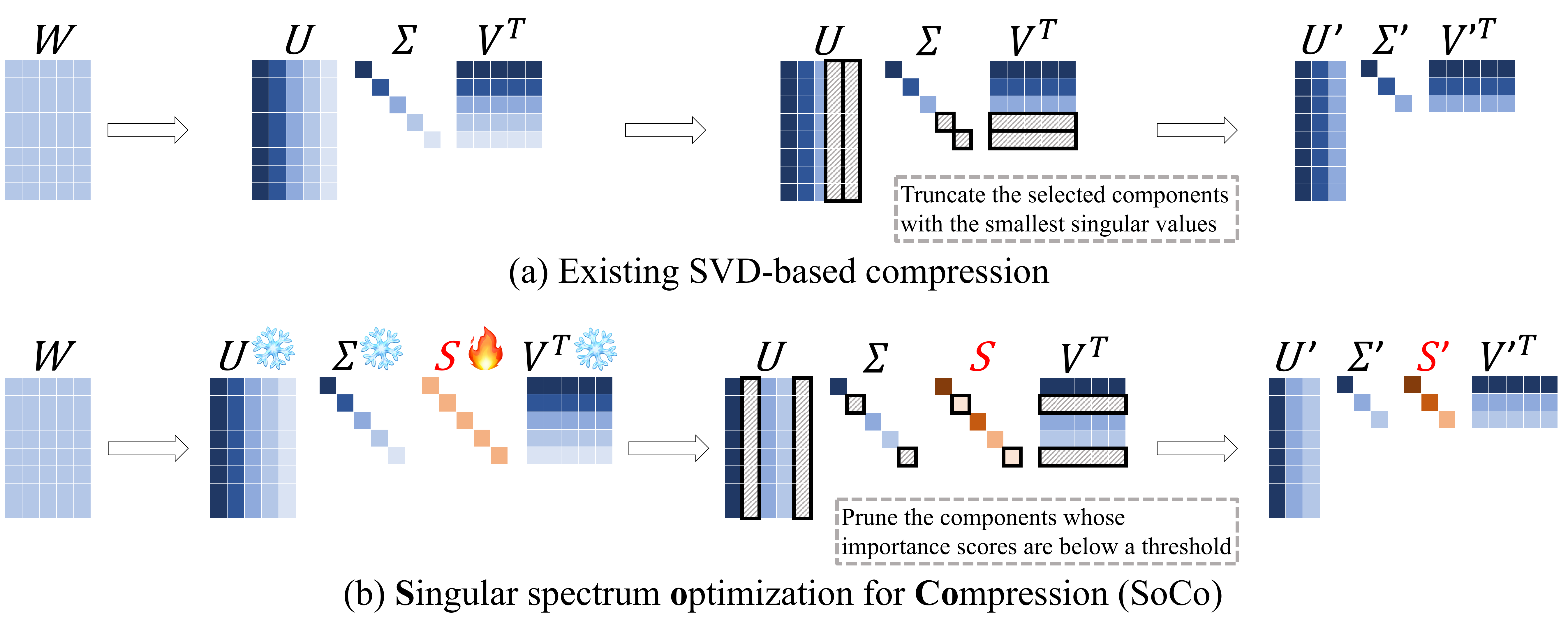}
\vspace{-8mm}
\caption{Illustration of the overall framework of SoCo. The pre-trained weight matrix $W$ is decomposed into $U \Sigma V^\top$, where $\Sigma$ is a diagonal matrix with diagonal elements arranged in descending order. a) Existing SVD-based compression methods truncate the smaller singular values and their corresponding vectors in $U$ and rows in $V^\top$. b) The proposed SoCo assigns a diagonal matrix $S$ as importance scores to singular values in $\Sigma$. 
% During training, only $S$ is updated, while the other parameters are frozen. 
After training, singular values with an importance score below a given threshold~(e.g., 0.5) are pruned. 
In particular, singular values with importance scores larger than the threshold rescale the preserved singular values to compensate the loss from pruned components.}
\label{fig:model}
\vskip -0.05in
\end{figure*}

\section{Method}
% In Section~\ref{sec:preliminary}, we briefly review LLM architectures and the use of SVD for model compression. Section~\ref{sec:Model Design} then details the SoCo~(\textbf{S}ingular spectrum \textbf{o}ptimization for large language model \textbf{Co}mpression) framework, while Section~\ref{sec:Training Strategy} describes our three-stage training strategy for refining importance scores and manipulating decomposed components. Finally, Section~\ref{sec:Analysis} analyzes the learning dynamics during SoCo’s optimization, highlighting how our method preserves essential model information while achieving effective parameter reduction.
In Section~\ref{sec:preliminary}, we review LLM architectures and SVD-based compression. Section~\ref{sec:Model Design} details the SoCo~(Singular spectrum optimization for large language model Compression) framework, while Section~\ref{sec:Training Strategy} outlines our three-stage training for refining importance scores and rescaling decomposed spectrum. Finally, Section~\ref{sec:Analysis} examines the learning dynamics during training, demonstrating how SoCo preserves essential model information while effectively reducing parameters.

\subsection{Preliminaries} \label{sec:preliminary}
Large language models are built on transformer layers that rely on Multi-Head Self-Attention~(MHSA) mechanisms. In each layer, the input sequence is transformed into queries, keys, and values via learned linear transformations to compute attention, followed by processing through a feed-forward network~(FFN). The weight matrices $W$~(used for query, key, value, and output projections in attention, as well as up, down, and gate projections in the FFN) comprise the majority of the model’s parameters. These high-dimensional matrices often exhibit redundancy, with key information concentrated in a few decomposed components~\cite{haink2023hessian, han2015deep}, making them the primary targets for compression.
% These matrices are high-dimensional and often exhibit redundancy, meaning that the essential information is concentrated in a few decomposed components. 
This phenomenon can be formally captured using SVD, which factorizes a weight matrix $W \in \mathbb{R}^{m \times n}$ as follows:
% \begin{small}
\begin{equation}
    W = U \Sigma V^\top,
\end{equation}
% \end{small} {\min(m,n)\times\min(m,n)}
where $\mathbb{R}$ denotes real numbers, $U \in \mathbb{R}^{m \times min(m,n)}$ and $V \in \mathbb{R}^{min(m,n) \times n}$ are orthogonal matrices and $\Sigma \in \mathbb{R}^{min(m,n) \times min(m,n)}$ is a diagonal matrix containing the singular values in descending order. The magnitude of each singular value quantifies the amount of energy in its corresponding direction, effectively identifying the principal decomposed components of the matrix~\cite{stewart1993early, wall2003singular}. By preserving only the top $k$ largest singular values and their associated vectors, SVD provides a low-rank approximation:
% \begin{small}
\begin{equation}
    W \approx U_k \Sigma_k V_k^\top,
\end{equation}
% \end{small}
which preserves the essential features of the parameter matrices. This property is useful for LLM compression, as it reduces LLM's storage and computational requirements when $k \ll \min (m,n)$.

\subsection{Overall Framework}\label{sec:Model Design}

Although recent approaches such as ASVD~\cite{yuan2023asvd} and SVD-LLM~\cite{wang2024svd} introduce compression-specific strategies, they still rely on truncating by the descending order of singular values from SVD. 
Since the original singular values only reflect their contributions to low-rank approximation, this rigid truncation is not necessarily in accordance with the true importance of each decomposed component for downstream tasks and may lead to suboptimal performance at higher compression ratios.

To address these limitations, we propose SoCo~(\textbf{S}ingular spectrum \textbf{o}ptimization for large language model \textbf{Co}mpression), a learning-based framework that adaptively re-evaluates and adjusts the singular spectrum. In SoCo, a trainable diagonal matrix $S \in \mathbb{R}^{min(m,n) \times min(m,n)}$ is introduced to assign importance scores to the singular values. Instead of selecting singular values based on their magnitude, the preserved singular values are determined according to the assigned scores in $S$, as illustrated in Fig.~\ref{fig:model}. The modified weight matrix $W^\prime$ is defined as
\begin{equation}
W^\prime = U ( \Sigma \odot S ) V^\top, \label{eq:svd_b}
\end{equation}
% \begin{equation}
% \tilde{\Sigma} = \Sigma \odot S,
% \end{equation}
where $\odot$ denotes element-wise multiplication. After training, $S$ dynamically adjusts the importance distribution of the decomposed components based on their true influences on model performance, thereby improving the trade-off between model compression and performance preservation.

We define the importance scores $S$ as:
\begin{equation}
S = \frac{\lambda_m}{1+e^{- \lambda_s z+\ln(2\lambda_m-1)}}, \label{eq:svd_a}
\end{equation}
where $z$ is a learnable diagonal matrix~(with the same shape as $\Sigma$), $\lambda_m$ controls the overall numerical scale of \(S\)~(thereby determining the magnitude of re-evaluation), and $\lambda_s$ modulates the steepness of this sigmoid-like function~(larger $\lambda_s$ results in a more rapid response to changes in $z$). The term $\ln(2\lambda_m-1)$ is employed to ensure that the function yields a consistent value at $z=0$.~(More details about $S$ are in ~\ref{sec:appendix_func} of Appendix.)

To compensate for any deviation introduced by modifying weight matrix $W'$, we further add a trainable deviation term $d$ after the linear transformation of $W'$.
% jointly optimized with $z$. 
The original model parameters, including the singular values $\Sigma$ and singular vectors $U$ and $V$, are frozen during our training. 

% During the training process, except $z$ and $d$, the others remain frozen.

\subsection{Three-stage Optimization}
\label{sec:Training Strategy}

\begin{figure}[ht]
\centerline{\includegraphics[width=\columnwidth]{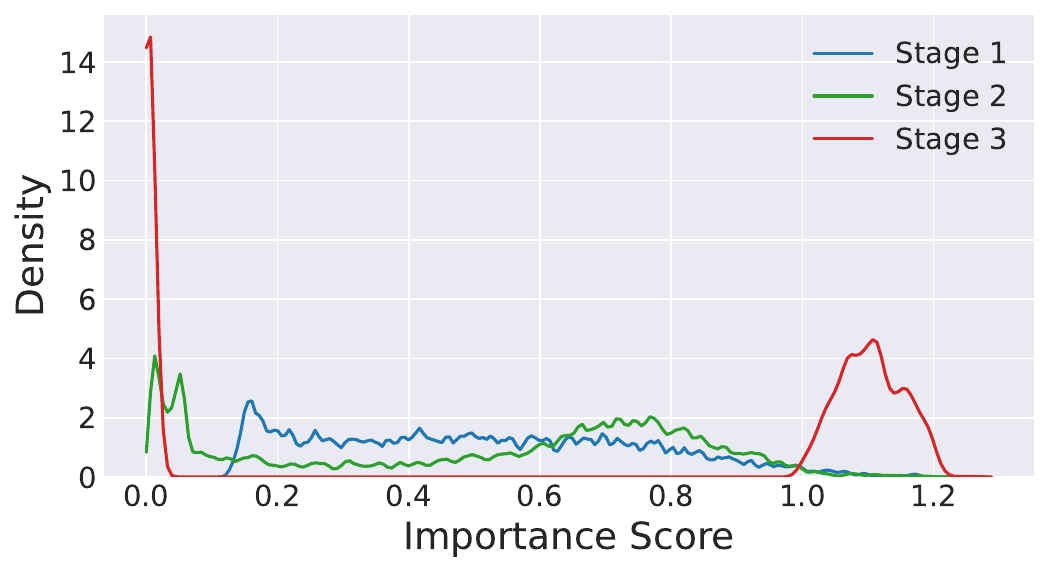}}
\vspace{-3mm}
\caption{Distribution of importance scores after each of our three training stages, illustrating the dynamic re-evaluation process.}
\vspace{-5mm}
\label{fig:dist3stages}
\end{figure}

Instead of a single-stage training approach, we propose a three-stage process to adaptively refine the importance scores assigned to the decomposed SVD components. Our approach is motivated by two key considerations. 1) Balancing conflict optimization objectives: compression and performance are inherently conflicting. Excessive parameter reduction can degrade performance while maintaining accuracy often requires retaining more parameters. 
Single-stage training must simultaneously optimize them, which is intractable to attain a satisfactory balance and may lead to suboptimal minima. 
%This tension leads to gradient interference, unstable optimization, and convergence to suboptimal minima. 
2) Empirical observation of importance score distributions: As shown in Fig.~\ref{fig:dist3stages}, the importance scores after Stage 1~(directly optimizing both compression and performance simultaneously) exhibit a relatively uniform and continuous distribution. 
As a result, pruning with a fixed threshold may remove many singular values near the threshold that still exert significant influence; when these values are pruned simultaneously across multiple layers, their cumulative effect can substantially degrade performance. Therefore, a more thorough evaluation of the importance of each singular value is needed, along with polarized and sparsified importance scores to prune components with minimal performance degradation.

In light of these considerations, our three-stage training process first rapidly achieves the compression ratio target in stage 1, then polarizes the importance score distribution to gradually separate essential components from redundant ones through alternating optimization in stage 2, and finally sparsifies the importance scores in stage 3. 
As shown in Fig.~\ref{fig:dist3stages}, after our three-stage training, there is a flat basin between the polarized scores, which minimizes the influence of using a proper threshold to prune the small scores.
The process is divided into three stages, each governed by a composite loss function. The corresponding conditions and loss functions are summarized in Tab.~\ref{tab:three_stages} and will be detailed in the following subsections.

\begin{table}%[htbp]
\centering
\caption{Conditions and loss functions for each training stage. $step$ denotes the procedure of the whole training process, where $step \in [0,1]$~(with 0 representing the start and 1 representing the end).}
\vspace{-3mm}
\begin{tabular}{ccc}
\toprule
Stage & Conditions & Loss Functions \\ \hline
1 & $ \makecell{step < 0.5 , r \geq R, \\ until \hspace{0.1cm} r < R} $ & $L_{inc} + L_{dec}$  \\ \hline
\multirow{2}{*}{2} & $ step < 0.5, r < R $ & $ L_{inc} $  \\ \cline{2-3} 
                  & $ step < 0.5 , r \geq R $ &  $L_{inc} + L_{dec}$  \\ \hline
3 &  $step \geq 0.5 $ & $L_{inc} + L_{spa}$ \\ \bottomrule
\end{tabular}
\label{tab:three_stages}
\end{table}

%\vspace{-5mm}
\subsubsection{Stage 1: Rapid and Rough Compression} \label{sec:stage1}
At the beginning of Stage 1, the current compression ratio $r=1.0$ exceeds the target compression ratio $R<1.0$. In this stage, both $L_{dec}$ and $L_{inc}$ jointly drive $r$ from 1.0 to $R$. The compression ratio $r$, which is also used as $L_{dec}$, is defined as:
\begin{equation}
\label{eq:l_decreasing}
L_{dec} = r =
\frac{
  \sum\limits_{l=1}^{L} \left(in_l+out_l\right) \times  c_l 
}{
  \sum\limits_{l=1}^{L} \left(in_l \times out_l\right)
}, 
\end{equation}
where \(in_l\) and \(out_l\) denote the input and output dimensions of layer \(l\), \(L\) denotes the number of transformer layers in certain LLM, and \(c_l\) is the number of selected decomposed components to preserve in the layer. When working as $L_{dec}$, to enable gradient flow through the hard thresholding operation, we use the Straight-Through Estimator~(STE)~\cite{bengio2013estimating, courbariaux2015binaryconnect}:
\begin{gather}
\begin{aligned}
c_l &= \sum\limits_{d=1}^{D}  \mathbf{1} \left(S^d_l \geq 0.5 \right) 
%\label{eq:l_decreasing_c_a}  
\\
&= \sum\limits_{d=1}^{D}  \mathbf{1} \left(S^d_l \geq 0.5 \right) - sg[S^d_l] + S^d_l, 
\label{eq:l_decreasing_c_b}
\end{aligned}
\end{gather}
where \(S^d_l\) is the importance score of the \(d\)-th singular value of layer \(l\) and \(sg[\cdot]\) halts gradient propagation. $\mathbf{1}(\cdot)$ represents the indicator function, which returns 1 if the condition is true and 0 otherwise. The loss \(L_{dec}\) thus applies a positive gradient to reduce \(S\), favoring compression. 
% Moreover, we encapsulate \(S\) to ensure that its gradient is consistently 1 or -1, with larger singular values receiving smaller gradients, thereby slowing their reduction~(See Algorithm~\ref{alg:BackSign} in Appendix for details).

Simultaneously, $L_{inc}$ counteracts excessive compression to maintain performance. We define
\begin{equation}
\label{eq:l_increasing} 
\mathcal L_{inc}  = \frac{1}{T} \sum\limits_{t=1}^{T} \left( \sum\limits_{c=1}^{C} y^c_t \log \frac{y^c_t}{p^c_t} \right),
\end{equation}
where \(T\) is the number of tokens at the current training step, \(C\) is the vocabulary size, \(y^c_t\) and \(p^c_t\) are the predicted probability distributions before and after model compression. This KL-divergence loss~\cite{Cobbe2021kl} forces the model to preserve performance during compressing. In practice, the gradient of $L_{inc}$ is smaller than $L_{dec}$'s gradient in Stage 1. Consequently, \(L_{dec}\) becomes the dominant factor, driving rapid model compression until \(r < R\). At that point, Stage 2 begins.

\subsubsection{Stage 2: Alternating Optimization}\label{sec:stage2}

After Stage 1, the distribution of importance scores in Fig.~\ref{fig:dist3stages} shows no clear cutoff for selecting critical components, rendering fixed-threshold pruning inefficient. In Stage 2, we employ an alternating optimization strategy: when \(r < R\), \(L_{dec}\) is inactive so that \(L_{inc}\) enables the recovery of critical components that are pruned too aggressively; when \(r \ge R\), \(L_{dec}\) is reactivated. This alternation induces controlled oscillations around the target ratio \(R\). With continued training and increased data exposure, the importance assessments gradually stabilize. Such re-evaluation leading to a better singular spectrum for compression~(see Tab.~\ref{tab:ablation_stages}).

\subsubsection{Stage 3: Importance Sparsity}\label{sec:stage3}

Following rapid convergence in Stage 1 and alternating optimization in Stage 2, enforcing sparsity in the importance scores becomes essential for clearly distinguishing essential components from redundant ones. In Stage 3, we introduce a sparsity loss $L_{spa}$ to drive $S^d_l \le 0.5$ toward 0~(to prune) and $ 0.5 < S^d_l \le 1.0$ toward 1~(to preserve):
\begin{equation}
\mathcal L_{spa}=\frac{1}{LD}  \sum\limits_{l=1}^{L} \sum\limits_{d=1}^{D} spa^d_l,
\label{eq:l_sparsity}
\end{equation}
where the quadratic penalty $spa^d_l$ is defined as
\vspace{-2mm}
\begin{equation}
spa^d_l = 
\begin{cases}
(S^d_l)^2, & \text{if } S^d_l \leqslant 0.5, \\
(S^d_l-1)^2, & \text{if } 0.5 < S^d_l \leqslant 1.0, \\
0, & \text{if } 1.0 < S^d_l.
\end{cases}
\label{eq:l_sparsity_spa}
\end{equation}

\noindent \(spa^d_l\) further drives \(S\) toward a bimodal distribution, as shown in Fig.~\ref{fig:dist3stages}. Combined with the continued optimization of \(L_{inc}\), Stage 3 reduces the contributions of pruned components while reinforcing essential components. This process creates the basin region of red curve in Fig.~\ref{fig:dist3stages} where an proper importance score threshold can be easily set to preserve performance with minimal accuracy degradation.

\subsection{Analysis}
\label{sec:Analysis}

\begin{figure}[t]
\centerline{\includegraphics[width=1.0\columnwidth]{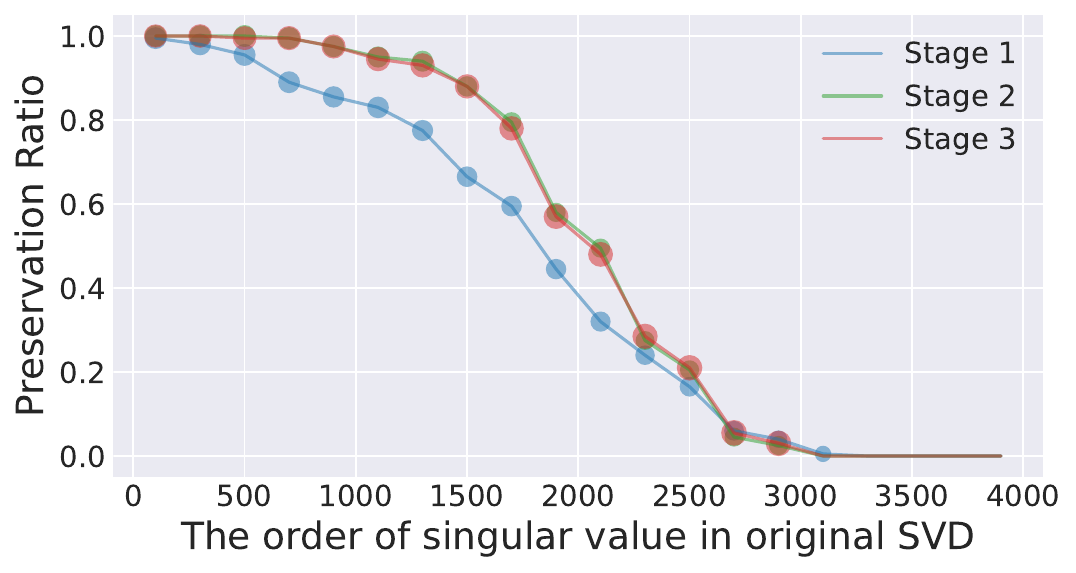}}
\vspace{-3mm}
\caption{Preservation ratio of components in the original SVD-based importance order across the three stages.}
\label{fig:keepingratio_3stages}
\end{figure}

As a pioneering work that integrates decomposed component re-evaluation into SVD-based compression, SoCo provides interesting insights. We summarize our findings as follows:

\noindent \textbf{Validation of Three-stage Optimization:} 
As shown in Tab.~\ref{tab:ablation_stages}, each training stage contributes to progressive performance gains, highlighting the effectiveness of phased optimization. This phenomenon validates our design propels the model beyond local optima to a more optimal state.

\noindent \textbf{Importance Score Distribution Evolution}  
Fig.~\ref{fig:dist3stages} illustrates the evolution of importance score distributions across training stages. In Stage 1, the scores are broadly and continuously distributed. In Stage 2, the distribution polarizes toward two opposing numerical extremes. By Stage 3, $S$ is driven further toward a bimodal distribution, with scores converging near 0.0~(discardable) and 1.0~(critical). This separation minimizes ambiguity and limits loss accumulation in threshold-based pruning, as fewer important components fall near the cutoff.

\noindent \textbf{Adaptation for Component Preservation:} 
Notably, the emergence of importance scores exceeding 1.0~(the right peak of red curve in Fig.~\ref{fig:dist3stages}) unveils a novel mechanism in our framework: the combination of sparsity and output-alignment loss in Stage 3 amplify preserved singular values to counterbalance the loss from pruned components. Unlike existing SVD-based compression methods, which truncate decomposed components, our SoCo can rescale the singular spectrum broadening the effectiveness and potential of SVD.

As shown in Fig.~\ref{fig:keepingratio_3stages}, the evolution of the top singular value preservation ratios across the three stages reveals the intrinsic adjustments in component importance. This dynamic re-assessment ensures the real importance distribution of decomposed components for model compression.

\section{Experiments}
\label{sec:experiments}

\subsection{Experiment Settings}

\noindent \textbf{Datasets:}  
% WikiText-2~\cite{Merity2016wiki} is partitioned training, validation, and testing sets. 
We use WikiText-2~\cite{Merity2016wiki} training set as the primary training corpus for our pipeline. For evaluation, we employ the WikiText-2~\cite{Merity2016wiki} and C4~\cite{Raffel2019C4}, and LM-Evaluation-Harness~\cite{eval-harness} testing sets to ensure a comprehensive assessment.
% of both linguistic fidelity and task-specific performance.

\noindent \textbf{Baselines:}  
Our experiments assess seven LLMs spanning five architectural families and three parameter scales: LLaMA-7B/13B/30B~\cite{touvron2023llama}, LLaMA2-7B~\cite{touvron2023llama2}, OPT-6.7B~\cite{zhang2022opt}, Vicuna-7B~\cite{vicuna2023}, and Mistral-7B~\cite{jiang2023mistral}.

\noindent \textbf{Evaluation Metrics:}  
\begin{itemize}[leftmargin=0.8em, itemsep=0.2ex, topsep=0.5ex, parsep=0ex]
    \item \textbf{Perplexity~(PPL):} on WikiText-2~\cite{Merity2016wiki} and C4~\cite{Raffel2019C4}.
    \item \textbf{Task-Specific Accuracy:} Evaluated using LM-Evaluation-Harness~\cite{eval-harness} on:
    \begin{itemize}[leftmargin=0.9em, itemsep=0.2ex, topsep=0.5ex, parsep=0ex]
        \item 6 classification tasks: OpenbookQA~\cite{Mihaylov2018OpenbookQA}, WinoGrande~\cite{Sakaguchi2019WinoGrande}, HellaSwag~\cite{Zellers2019HellaSwag}, ARC-e~\cite{Clark2018Arc}, PIQA~\cite{Bisk2019PIQARA}, and MathQA~\cite{Amini2019MathQATI}. ``Average$\uparrow$'' in the tables below denotes the mean accuracy across the six classification tasks.
        \item 1 generative task: TruthfulQA~(factual consistency)~\cite{Lin2021TruthfulQAMH}.
    \end{itemize}
\end{itemize}

\begin{table*}[ht]
% \vskip 0.15in
\caption{Comparison of SoCo with other SVD-based compression methods on the LLaMA-7B model across various compression ratios. The training dataset is the WikiText-2 training set, and evaluation is conducted on the WikiText-2, C4, and LM-Evaluation-Harness testing sets. 
% ``Average$\uparrow$'' in the tables below denotes the mean accuracy across six classification tasks on LM-Evaluation-Harness.
}
\label{tab:svd_methods}
\vspace{-0.19in}
\begin{center}
\resizebox{1\textwidth}{!}{
\begin{tabular}{l|l|cc|ccccccc|c}
% \rowcolors{1}{}{lightgray}
% \begin{tabular}{llcccccccccc}
\toprule
\textsc{Ratio} & \textsc{Method} & WikiText-2$\downarrow$ & C4$\downarrow$ & Openb.$\uparrow$ & ARC\_e$\uparrow$ & WinoG.$\uparrow$ & HellaS.$\uparrow$ & PIQA$\uparrow$ & MathQA$\uparrow$ & Average$\uparrow$ & TruQA.$\uparrow$ \\
\hline
0\%  & Original      & 5.68           & 7.34           & 34\%          & 75\%          & 70\%          & 57\%          & 79\%          & 27\%          & 57\%          & 30\%                   \\ \hline
20\% & SVD           & 20061          & 18800          &  5\%          &  4\%          &  1\%          &  3\%          &  2\%          &  3\%          &  3\%          &  0\%                   \\
     & FWSVD         & 1727           & 1511           &  9\%          & 11\%          &  5\%          &  8\%          & 10\%          &  5\%          &  8\%          &  0\%                   \\
     & ASVD          & 11.14          & 15.93          & 29\%          & 53\%          & 64\%          & 41\%          & 68\%          & 17\%          & 45\%          & 21\%                   \\
     & SVD-LLM       & 7.94           & 15.84          & \textbf{31\%} & 62\%          & 61\%          & 45\%          & 71\%          & 21\%          & 49\%          & 26\%                   \\
     & \textbf{SoCo} & \textbf{6.67}  & \textbf{9.89}  & 28\%          & \textbf{70\%} & \textbf{65\%} & \textbf{51\%} & \textbf{75\%} & \textbf{25\%} & \textbf{52\%} & \textbf{33\%}         \\ \hline
40\% & SVD           & 52489          & 47774          &  4\%          &  4\%          &  5\%          &  1\%          &  3\%          &  2\%          &  3\%          &  0\%                   \\
     & FWSVD         & 18156          & 12847          &  6\%          &  5\%          &  2\%          &  0\%          &  5\%          &  3\%          &  4\%          &  0\%                   \\
     & ASVD          & 1407           & 1109           &  8\%          & 11\%          &  9\%          &  8\%          & 13\%          &  8\%          & 10\%          &  1\%                   \\
     & SVD-LLM       & 13.73          & 75.42          & \textbf{25\%} & 33\%          & \textbf{61\%} & 40\%          & 63\%          & 12\%          & 39\%          & 17\%                   \\
     & \textbf{SoCo} & \textbf{8.60}  & \textbf{13.71} & 23\%          & \textbf{63\%} & 60\%          & \textbf{45\%} & \textbf{70\%} & \textbf{23\%} & \textbf{47\%} & \textbf{36\%}           \\ \hline
60\% & SVD           & 105474         & 106976         &  1\%          &  3\%          &  1\%          &  0\%          &  1\%          &  2\%          &  1\%          &  0\%                   \\
     & FWSVD         & 32194          & 29292          &  6\%          &  2\%          &  1\%          &  1\%          &  2\%          &  3\%          &  3\%          &  0\%                   \\
     & ASVD          & 57057          & 43036          &  5\%          &  4\%          &  6\%          &  9\%          &  8\%          &  5\%          &  6\%          &  0\%                   \\
     & SVD-LLM       & 66.62          & 471.83         & 10\%          &  5\%          & 17\%          & 10\%          & 21\%          &  4\%          & 11\%          &  1\%                   \\
     & \textbf{SoCo} & \textbf{12.20} & \textbf{24.53} & \textbf{19\%} & \textbf{48\%} & \textbf{55\%} & \textbf{35\%} & \textbf{63\%} & \textbf{21\%} & \textbf{40\%} & \textbf{39\%}           \\ \hline
80\% & SVD           & 687291         & 708243         &  0\%          &  4\%          &  2\%          &  1\%          &  1\%          &  0\%          &  1\%          &  0\%                   \\
     & FWSVD         & 96872          & 89243          &  1\%          &  2\%          &  0\%          &  6\%          &  9\%          &  0\%          &  3\%          &  0\%                   \\
     & ASVD          & 80425          & 67927          &  4\%          &  3\%          &  3\%          &  7\%          & 10\%          &  1\%          &  5\%          &  0\%                   \\
     & SVD-LLM       & 1349           & 6224           &  7\%          &  1\%          & 12\%          & 10\%          &  7\%          &  6\%          &  7\%          &  0\%                   \\
     & \textbf{SoCo} & \textbf{21.03} & \textbf{50.86} & \textbf{14\%} & \textbf{36\%} & \textbf{53\%} & \textbf{28\%} & \textbf{57\%} & \textbf{20\%} & \textbf{35\%} & \textbf{ 0\%}           \\
\bottomrule
\end{tabular}
}
\end{center}
\end{table*}

% \vspace{-0.6in}

\begin{table*}[ht]
%在不同家族系列和不同大小的模型上，在压缩率为20\%时，与其他低秩分解方法相比较。在WikiText-2测试集上测试PPL指标。使用LM-Evaluation-Harness framework to evaluate performance on six classification datasets. 所有实验在模型压缩之后，都未经过微调。
\begin{center}
%\vspace{-0.15in}
\caption{Comparison of SoCo with other SVD-based compression methods on various model families at 20\% compression ratio. The training dataset is WikiText-2, while evaluation is conducted on the WikiText-2 and LM-Evaluation-Harness's 6 classification task testing sets. 
% ``Average$\uparrow$'' in the tables below denotes the mean accuracy across six classification tasks on LM-Evaluation-Harness.
}
\label{tab:other_model_families}
\vspace{-0.14in}
\resizebox{1\textwidth}{!}{
% \rowcolors{1}{}{lightgray}
% \begin{tabular}{l|cc|cc|cc|cc|cc|cc}
\begin{tabular}{lcccccccccccc}
\toprule
         & \multicolumn{2}{c}{OPT-6.7B} & \multicolumn{2}{c}{LLaMA2-7B} & \multicolumn{2}{c}{Mistral-7B} & \multicolumn{2}{c}{Vicuna-7B} & \multicolumn{2}{c}{LLaMA-13B} & \multicolumn{2}{c}{LLaMA-30B} \\
\hline
\textsc{Method}   & WikiText-2$\downarrow$ & Average$\uparrow$ & WikiText-2$\downarrow$ & Average$\uparrow$ & WikiText-2$\downarrow$ & Average$\uparrow$ & WikiText-2$\downarrow$ & Average$\uparrow$ & WikiText-2$\downarrow$ & Average$\uparrow$ & WikiText-2$\downarrow$ & Average$\uparrow$ \\
\hline
Original      & 10.86          & 52\%          & 5.47          & 57\%          & 5.25          & 61\%          & 6.78          & 56\%          & 5.09          & 59\%          & 4.10          & 61\%          \\
\hline
SVD           & 66275          &  3\%          & 18192         &  9\%          & 159627        &  3\%          & 18644         &  5\%          & 946.31        & 21\%          & 54.11         & 33\%          \\
FWSVD         & 14559          &  6\%          & 2360          & 12\%          & 6357          &  8\%          & 2758          &  9\%          & 15.98         & 43\%          & 20.54         & 42\%          \\
ASVD          & 82.00          & 32\%          & 10.10         & 36\%          & 13.72         & 32\%          & 16.23         & 33\%          & 6.74          & 54\%          & 22.71         & 44\%          \\
SVD-LLM       & 16.04          & 41\%          & 8.50          & \textbf{53\%} & 10.21         & 42\%          & 8.41          & 51\%          & 6.61          & 54\%          & 5.63          & 57\%          \\
 \textbf{SoCo} & \textbf{10.44} & \textbf{50\%} & \textbf{6.69} & 52\%          & \textbf{6.69} & \textbf{54\%} & \textbf{7.34} & \textbf{53\%} & \textbf{5.92} & \textbf{56\%} & \textbf{5.17} & \textbf{58\%} \\
\bottomrule
\end{tabular}
}
% \vspace{-5mm}
\end{center}
\end{table*}

\subsection{Main Results}

For a comprehensive comparison, we evaluate SoCo against two categories of model compression techniques: SVD-based methods and pruning methods. For fair evaluations, all experiments are conducted without fine-tuning after compression.

\subsubsection{Comparison with SVD-based Methods}
% \vspace{-2mm}
In Tab.~\ref{tab:svd_methods}, SoCo consistently outperforms the competing SVD-based approaches~\cite{Hsu2022FWSVD, yuan2023asvd, wang2024svd} across compression ratios of 20\%, 40\%, 60\%, and 80\%. Notably, compared with SVD-LLM~\cite{wang2024svd}, SoCo achieves relative performance improvements of up to 37.6\%, 81.8\%, 94.8\%, and 99.2\% on the C4 dataset~\cite{Raffel2019C4}, corresponding to absolute performance gains ranging from 15.84 to 9.89, 75.42 to 13.71, 471.83 to 24.53, and 6224 to 50.86, respectively. Especially at compression ratios of 60\% and 80\%, where previous SVD-based methods become impractical, SoCo is the only approach capable of delivering robust performance. 
Next, Tab.~\ref{tab:other_model_families} compares our SoCo with other SVD-based compression approaches~\cite{Hsu2022FWSVD, yuan2023asvd, wang2024svd} across different model families and sizes at 20\% compression ratio. 
SoCo consistently achieves superior performance, 
demonstrating its general applicability across diverse model architectures and sizes.

\subsubsection{Comparison with Pruning Methods}

We compare SoCo with pruning methods, including LLM-Pruner~\cite{ma2023llm}, SliceGPT~\cite{ashkboos2024slicegpt}, and BlockPruner~\cite{zhong2024blockpruner}, on the LLaMA-7B~\cite{touvron2023llama} at compression ratios of 26\%, 33\%, 40\%, and 47\%. 
For fair comparison, we follow the baseline methods using the Alpaca~\cite{taori2023alpaca} dataset in training, and performance is evaluated using PPL on the WikiText-2 testing set in Tab.~\ref{tab:pruning_methods}.
SoCo consistently outperforms these pruning methods.

\subsection{Ablation Study} \label{Ablation Study}

\begin{table}[ht]
%在LLaMA-7B模型上，在压缩率为20\%时，与剪枝方法相比较。训练数据集为Alpaca数据集，在WikiText-2测试集上测试zero-shot的PPL指标。
\caption{Comparison with pruning methods on the LLaMA-7B model at compression ratios of 26\%, 33\%, 40\%, and 47\%. Following the compared methods, we use the Alpaca dataset for training, and the PPL metric is tested on the WikiText-2 testing set.}
\vspace{-0.18in}
\begin{center}
\resizebox{0.50\textwidth}{!}{
% \rowcolors{1}{}{lightgray}
\begin{tabular}{lcccc}
\toprule
\textsc{Method}  & 10GB(26\%) & 9GB(33\%)   & 8GB(40\%)   & 7GB(47\%)   \\
\hline
LLM-Pruner       & 9.88 & 12.21 & 18.94 & 21.68 \\
SliceGPT        & 8.78 & 12.73 & 16.39 & 27.41 \\
BlockPruner      & 9.40  & 12.76 & 19.78 & 43.05 \\
%\textbf{SoCo} & 7.25 & 7.53  & 8.06  & 9.04  \\
%\textbf{SoCo$_{alpaca}$} & 8.44 & 9.49  & 11.06  & 14.08  \\
%\textbf{SoCo$_{alpaca}$} & \textbf{8.14} & \textbf{9.21}  & \textbf{11.32}  & \textbf{13.21}  \\
 \textbf{SoCo$_{alpaca}$} & \textbf{8.67} & \textbf{10.13}  & \textbf{12.36}  & \textbf{16.50}  \\
\bottomrule
\end{tabular}
}
\end{center}
\label{tab:pruning_methods}
% \vspace{-2mm}
\end{table}

%We use the LLaMA-7B model with a compression ratio of 20\% to ablate the effect of the designs in SoCo below.

We conduct an ablation study on LLaMA-7B~\cite{touvron2023llama}.
% at 20\% compression ratio.

% To evaluate the contributions of each design in our SoCo framework, we conduct the ablation study on the LLaMA-7B~\cite{touvron2023llama} at 20\% compression ratio. 

\noindent \textbf{Effect of Three-Stage Optimization.}  
Tab.~\ref{tab:ablation_stages} reports the performance on the WikiText-2~\cite{Merity2016wiki} and C4~\cite{Raffel2019C4}, and LM-Evaluation-Harness~\cite{eval-harness} testing sets.
The results demonstrate that our three-stage process progressively refines the importance scores, consistently improving performance. The full three-stage SoCo achieves the best trade-off between compression and performance across all compression ratios.

\begin{table}[t]
\caption{Ablation study of the three-stage training process on the LLaMA-7B~\cite{touvron2023llama} model at compression ratio of 20\%. Performance is measured using  PPL on WikiText-2 and C4, and the average accuracy on LM-Evaluation-Harness’s 6 classification tasks.
}
\vspace{-0.18in}
\begin{center}
\resizebox{0.50\textwidth}{!}{
% \rowcolors{1}{}{lightgray}
\begin{tabular}{clccc}
\toprule
\textsc{Ratio} & \textsc{Stage} & WikiText-2$\downarrow$ & C4$\downarrow$   & Average$\uparrow$  \\
\hline
20\% & 1                    & 16.21          & 24.72          & 41.41\%          \\
     & 1+2                  & 7.13           & 10.76          & 50.88\%          \\
     & \textbf{1+2+3(SoCo)} & \textbf{6.67}  & \textbf{9.89}  & \textbf{52.48\%} \\
\hline
40\% & 1                    & 109.80         & 178.16         & 33.54\%          \\
     & 1+2                  & 27.05          & 57.14          & 41.21\%          \\
     & \textbf{1+2+3(SoCo)} & \textbf{8.60}  & \textbf{13.71} & \textbf{47.16\%} \\
\hline
60\% & 1                    & 77.51          & 146.26         & 32.32\%          \\
     & 1+2                  & 31.26          & 66.29          & 33.06\%          \\
     & \textbf{1+2+3(SoCo)} & \textbf{12.20} & \textbf{24.53} & \textbf{40.22\%} \\
\hline
80\% & 1                    & 69.86          & 165.45         & 32.68\%          \\
     & 1+2                  & 44.37          & 104.46         & 33.41\%          \\
     & \textbf{1+2+3(SoCo)} & \textbf{21.03} & \textbf{50.86} & \textbf{34.63\%} \\
\bottomrule
\end{tabular}
}
\end{center}
\vspace{-2mm}
\label{tab:ablation_stages}
\end{table}

\noindent \textbf{Training Dataset Analysis.}  
To illustrate the task-aware effect on compression performance, we evaluate our method using four training datasets: the first 20.48 million tokens of the C4~\cite{Raffel2019C4} set, the Alpaca~\cite{taori2023alpaca} training dataset, the WikiText-2~\cite{Merity2016wiki} training set, and the PTB~\cite{marcus1993building} training set. Tab.~\ref{tab:ablation_datasets} shows that domain-specific training leads to superior performance on its corresponding test set; for example, when using the C4 set for learning the singular spectrum, the compressed model performs best on the C4 test set. These results indicate that task-aware compression further enhances performance. Our SoCo allows for task-aware singular spectrum rescaling when optimal domain performance is desired. In practice, however, the pruned model may not target a specific domain or may lack dedicated training data. Thus, we use WikiText-2~\cite{Merity2016wiki} as the default training set in our general-setting experiments (Tab.~\ref{tab:svd_methods} and Tab.~\ref{tab:other_model_families}). 

%all other experiments in this paper are conducted using a single training dataset (e.g., WikiText-2) across various datasets and test settings.
% \textcolor{red}{task aware, jin yi bu ti sheng xinneng. sometimes there is no data on target domain}
% \textcolor{red}{but for a general compression}
% To illustrate the task-aware effect of training data on compression performance, we evaluate the compressed LLaMA-7B model (at a 20\% compression ratio) using three distinct training datasets: the full Alpaca dataset, the complete WikiText-2 training set, and the first 20.48 million tokens of the C4 training set. Tab.~\ref{tab:ablation_datasets} shows that training on a domain-specific dataset leads to superior performance on its corresponding test set. For example, when C4 is used for training, the model performs best on the C4 test set; similarly, training on WikiText-2 results in improved performance on the WikiText-2 test set. These results underscore the benefit of task-aware training for achieving optimal model performance.

\begin{table}[t]
\caption{Ablation study on domain-specific training datasets for the LLaMA-7B model at 20\% compression ratio. The training datasets include the first 20.48M tokens of the C4 training set, the PTB training set, the Alpaca training set, and the WikiText-2 training set. Evaluation is performed on the C4, PTB, Alpaca, WikiText-2 and LM-Evaluation-Harness test sets.
}
\vspace{-0.18in}
\begin{center}
\resizebox{0.50\textwidth}{!}{
% \rowcolors{1}{}{lightgray}
\begin{tabular}{lccccc}
\toprule
\textsc{Dataset}     & C4$\downarrow$ & PTB$\downarrow$  & Alpaca$\downarrow$ & WikiText-2$\downarrow$ & Average$\uparrow$ \\
\hline
C4                   & \textbf{9.02} & 13.70 & 4.76  & 7.22 & 52.25\% \\
PTB                  & 11.93 & \textbf{12.63} & 5.05 & 8.77 & 51.64\% \\
Alpaca               & 9.91 & 16.17 & \textbf{4.61}  & 7.70 & 52.12\%\\
WikiText-2           & 9.89 & 13.61 & 4.90 & \textbf{6.67} & \textbf{52.48\%}\\
\bottomrule
\end{tabular}
}
\end{center}
\label{tab:ablation_datasets}
\vspace{-1mm}
\end{table}

% \begin{table}[ht]
% \caption{Ablation study on domain-specific training datasets for the LLaMA-7B model at a 20\% compression ratio. The training datasets include the full Alpaca dataset, the complete WikiText-2 training set, and the first 20.48M tokens of the C4 training set. Evaluation is performed on the WikiText-2, C4, and LM-Evaluation-Harness test sets.
% }
% \vspace{-0.2in}
% \begin{center}
% \resizebox{0.48\textwidth}{!}{
% \rowcolors{1}{}{lightgray}
% \begin{tabular}{lcccccc}
% \toprule
% \textsc{Dataset}     &Token no. & C4$\downarrow$ & PTB$\downarrow$  & Alpaca$\downarrow$ & WikiText-2$\downarrow$ & Average$\uparrow$ \\
% \hline
% C4                         & 20.48M & \textbf{8.99} & 13.96 & 4.75  & 7.18 & 52.25\% \\
% PTB                        & 1.24M  & 11.93 & \textbf{12.63} & 5.05 & 8.77 & 51.64\% \\
% Alpaca                     & 9.57M  & 9.91 & 16.17 & \textbf{4.61}  & 7.70 & 52.12\%\\
% WikiText-2                 & 2.87M  & 9.89 & 13.61 & 4.90 & \textbf{6.67} & \textbf{52.48\%}\\
% \bottomrule
% \end{tabular}
% }
% \end{center}
% \label{tab:ablation_datasets}
% \end{table}

\begin{table}[t]
%在不同大小的模型上时，可训练的参数量、所消耗的资源和所需的训练时间。
\caption{Trainable parameters, GPU resources, and training time required for different model sizes.}
\vspace{-0.18in}
\begin{center}
\resizebox{0.50\textwidth}{!}{
% \rowcolors{1}{}{lightgray}
\begin{tabular}{lccc}
\toprule
\textsc{Resource}          & LLAMA-7B                        & LLAMA-13B                       & LLAMA-30B                       \\
\hline
Trainable parameters & 1.84M (2.83\textpertenthousand) & 2.87M (2.26\textpertenthousand) & 5.59M (1.74\textpertenthousand) \\
GPU number           & 1                               & 1                               & 4                               \\
Memory per GPU       & 23G                             & 44G                             & 36G                             \\
Training time        & 2h10m                         & 3h10m                         & 2h45m                         \\
\bottomrule
\end{tabular}
}
\end{center}
\label{tab:resource}
\end{table}

\noindent \textbf{Training Time and GPU Cost.}
Tab.~\ref{tab:resource} summarizes the trainable parameter counts, GPU requirements, and training durations across various model sizes. Our results indicate that SoCo is highly resource-efficient, requiring only a small number of trainable parameters and enabling rapid singular spectrum optimization with modest GPU resources.

\section{Conclusion}
% \noindent \textbf{Limitation.} Our method currently employs a fixed threshold~(e.g., 0.5) for pruning based on the importance scores $S$. Although this threshold is effective in our experiments, it may not fully capture the nuanced variations in component importance across different model architectures or tasks. Future work could explore adaptive thresholding strategies to further enhance compression performance without compromising general applicability.

% In this paper, we introduce SoCo (\textbf{S}ingular spectrum \textbf{o}ptimization for large language model \textbf{Co}mpression), a novel framework that uses a learnable diagonal matrix to adaptively rescale the singular spectrum of weight matrices. Unlike traditional SVD truncation methods, SoCo re-evaluates the intrinsic importance of decomposed components and prunes them accordingly. Our three-stage training process progressively refines these scores, enabling a more effective compression. Experiments on multiple large language models and benchmarks demonstrate that SoCo outperforms existing SVD-based and pruning methods, achieving superior performance even at high compression ratios.

In this paper, we introduce SoCo (\textbf{S}ingular spectrum \textbf{o}ptimization for large language model \textbf{Co}mpression), a novel compression framework that leverages a learnable diagonal matrix to adaptively rescale the singular spectrum of weight matrices. Unlike traditional SVD-based truncation methods, SoCo re-evaluates the intrinsic importance of decomposed components and prunes them accordingly. Our three-stage training process progressively refines the importance scores of the decomposed SVD components, enabling a more fine-grained and effective compression strategy. Extensive experiments on multiple large language models and benchmarks demonstrate that SoCo consistently outperforms existing SVD-based and pruning methods, delivering superior performance even at high compression ratios. These results highlight the potential of our adaptive, data-driven approach for model compression, paving the way for more efficient deployment of large language models in resource-constrained environments.

\section{Limitations}
Our method currently employs a fixed threshold~(e.g., 0.5) for pruning based on the importance scores $S$. Although this threshold is effective in our experiments, it may not fully capture the nuanced variations in component importance across different model architectures or tasks. Future work could explore adaptive thresholding strategies to further enhance compression performance without compromising general applicability.

% \newpage
\bibliography{acl2025_sso}

@article{haink2023hessian,
  title={Hessian Eigenvectors and Principal Component Analysis of Neural Network Weight Matrices},
  author={Haink, David},
  journal={arXiv preprint arXiv:2311.00452},
  year={2023}
}

@article{han2015deep,
  title={Deep compression: Compressing deep neural networks with pruning, trained quantization and huffman coding},
  author={Han, Song and Mao, Huizi and Dally, William J},
  journal={arXiv preprint arXiv:1510.00149},
  year={2015}
}

@inproceedings{waswani2017attention,
  title={Attention is all you need},
  author={Waswani, A and Shazeer, N and Parmar, N and Uszkoreit, J and Jones, L and Gomez, A and Kaiser, L and Polosukhin, I},
  booktitle={NeurIPS},
  year={2017}
}

@article{brown2020Language,
  title={Language models are few-shot learners},
  author={Tom B. Brown and Benjamin Mann and Nick Ryder and Melanie Subbiah and Jared Kaplan and Prafulla Dhariwal and Arvind Neelakantan and Pranav Shyam and Girish Sastry and Amanda Askell and Sandhini Agarwal and Ariel Herbert-Voss and Gretchen Krueger and Tom Henighan and Rewon Child and Aditya Ramesh and Daniel M. Ziegler and Jeff Wu and Clemens Winter and Christopher Hesse and Mark Chen and Eric Sigler and Ma-teusz Litwin and Scott Gray and Benjamin Chess and Jack Clark and Christopher Berner and Sam McCandlish and Alec Radford and Ilya Sutskever and Dario Amodei},
  journal={NeurIPS},
  year={2020},
}

@article{wang2024svd,
  title={Svd-llm: Truncation-aware singular value decomposition for large language model compression},
  author={Wang, Xin and Zheng, Yu and Wan, Zhongwei and Zhang, Mi},
  journal={ICLR},
  year={2024}
}

@article{yuan2023asvd,
  title={Asvd: Activation-aware singular value decomposition for compressing large language models},
  author={Yuan, Zhihang and Shang, Yuzhang and Song, Yue and Wu, Qiang and Yan, Yan and Sun, Guangyu},
  journal={arXiv preprint arXiv:2312.05821},
  year={2023}
}

@article{Hsu2022FWSVD,
  title={Language model compression with weighted low-rank factorization},
  author={Yen-Chang Hsu and Ting Hua and Sung-En Chang and Qiang Lou and Yilin Shen and Hongxia Jin},
  journal={ICLR},
  year={2022},
}

@article{bengio2013estimating,
  title={Estimating or propagating gradients through stochastic neurons for conditional computation},
  author={Bengio, Yoshua and L{\'e}onard, Nicholas and Courville, Aaron},
  journal={arXiv preprint arXiv:1308.3432},
  year={2013}
}

@article{courbariaux2015binaryconnect,
  title={Binaryconnect: Training deep neural networks with binary weights during propagations},
  author={Courbariaux, Matthieu and Bengio, Yoshua and David, Jean-Pierre},
  journal={NeurIPS},
  year={2015}
}

@article{dettmers2022gpt3,
  title={Gpt3. int8 (): 8-bit matrix multiplication for transformers at scale},
  author={Dettmers, Tim and Lewis, Mike and Belkada, Younes and Zettlemoyer, Luke},
  journal={NeurIPS},
  year={2022}
}

@inproceedings{kimsqueezellm,
  title={SqueezeLLM: Dense-and-Sparse Quantization},
  author={Kim, Sehoon and Hooper, Coleman Richard Charles and Gholami, Amir and Dong, Zhen and Li, Xiuyu and Shen, Sheng and Mahoney, Michael W and Keutzer, Kurt},
  booktitle={ICML},
  year={2024}
}

@article{shen2024efficient,
  title={Efficient post-training quantization with fp8 formats},
  author={Shen, Haihao and Mellempudi, Naveen and He, Xin and Gao, Qun and Wang, Chang and Wang, Mengni},
  journal={MLSys},
  year={2024}
}

@article{ma2023llm,
  title={Llm-pruner: On the structural pruning of large language models},
  author={Ma, Xinyin and Fang, Gongfan and Wang, Xinchao},
  journal={NeurIPS},
  year={2023}
}

@article{ashkboos2024slicegpt,
  title={Slicegpt: Compress large language models by deleting rows and columns},
  author={Ashkboos, Saleh and Croci, Maximilian L and Nascimento, Marcelo Gennari do and Hoefler, Torsten and Hensman, James},
  journal={ICLR},
  year={2024}
}

@article{zhong2024blockpruner,
  title={BlockPruner: Fine-grained Pruning for Large Language Models},
  author={Zhong, Longguang and Wan, Fanqi and Chen, Ruijun and Quan, Xiaojun and Li, Liangzhi},
  journal={arXiv preprint arXiv:2406.10594},
  year={2024}
}

@article{zhong2024revisiting,
  title={Revisiting knowledge distillation for autoregressive language models},
  author={Zhong, Qihuang and Ding, Liang and Shen, Li and Liu, Juhua and Du, Bo and Tao, Dacheng},
  journal={ACL},
  year={2024}
}

@inproceedings{muralidharan2024compact,
  title={Compact language models via pruning and knowledge distillation},
  author={Muralidharan, Saurav and Sreenivas, Sharath Turuvekere and Joshi, Raviraj Bhuminand and Chochowski, Marcin and Patwary, Mostofa and Shoeybi, Mohammad and Catanzaro, Bryan and Kautz, Jan and Molchanov, Pavlo},
  booktitle={NeurIPS},
  year={2024}
}

@article{hsieh2023distilling,
  title={Distilling step-by-step! outperforming larger language models with less training data and smaller model sizes},
  author={Hsieh, Cheng-Yu and Li, Chun-Liang and Yeh, Chih-Kuan and Nakhost, Hootan and Fujii, Yasuhisa and Ratner, Alexander and Krishna, Ranjay and Lee, Chen-Yu and Pfister, Tomas},
  journal={ACL},
  year={2023}
}

@article{wan2023efficient,
  title={Efficient large language models: A survey},
  author={Wan, Zhongwei and Wang, Xin and Liu, Che and Alam, Samiul and Zheng, Yu and Liu, Jiachen and Qu, Zhongnan and Yan, Shen and Zhu, Yi and Zhang, Quanlu and others},
  journal={arXiv preprint arXiv:2312.03863},
  year={2023}
}

@article{zhu2023survey,
  title={A survey on model compression for large language models},
  author={Zhu, Xunyu and Li, Jian and Liu, Yong and Ma, Can and Wang, Weiping},
  journal={arXiv preprint arXiv:2308.07633},
  year={2023}
}

@article{stewart1993early,
  title={On the early history of the singular value decomposition},
  author={Stewart, Gilbert W},
  journal={SIAM review},
  year={1993},
}

@incollection{wall2003singular,
  title={Singular value decomposition and principal component analysis},
  author={Wall, Michael E and Rechtsteiner, Andreas and Rocha, Luis M},
  booktitle={A practical approach to microarray data analysis},
  pages={91--109},
  year={2003},
  publisher={Springer}
}

@article{Raffel2019C4,
  title={Exploring the Limits of Transfer Learning with a Unified Text-to-Text Transformer},
  author={Colin Raffel and Noam M. Shazeer and Adam Roberts and Katherine Lee and Sharan Narang and Michael Matena and Yanqi Zhou and Wei Li and Peter J. Liu},
  journal={JMLR},
  year={2019},
}

@article{Merity2016wiki,
  title={Pointer Sentinel Mixture Models},
  author={Stephen Merity and Caiming Xiong and James Bradbury and Richard Socher},
  journal={ICLR},
  year={2016},
}

@inproceedings{Mihaylov2018OpenbookQA,
  title={Can a Suit of Armor Conduct Electricity? A New Dataset for Open Book Question Answering},
  author={Todor Mihaylov and Peter Clark and Tushar Khot and Ashish Sabharwal},
  booktitle={EMNLP},
  year={2018},
}

@inproceedings{Sakaguchi2019WinoGrande,
  title={An Adversarial Winograd Schema Challenge at Scale},
  author={Keisuke Sakaguchi and Ronan Le Bras and Chandra Bhagavatula and Yejin Choi},
  year={2019},
  booktitle={AAAI},
}

@inproceedings{Zellers2019HellaSwag,
  title={HellaSwag: Can a Machine Really Finish Your Sentence?},
  author={Rowan Zellers and Ari Holtzman and Yonatan Bisk and Ali Farhadi and Yejin Choi},
  booktitle={ACL},
  year={2019},
}

@inproceedings{Bisk2019PIQARA,
  title={PIQA: Reasoning about Physical Commonsense in Natural Language},
  author={Yonatan Bisk and Rowan Zellers and Ronan Le Bras and Jianfeng Gao and Yejin Choi},
  booktitle={AAAI},
  year={2019},
}

@inproceedings{Amini2019MathQATI,
  title={MathQA: Towards Interpretable Math Word Problem Solving with Operation-Based Formalisms},
  author={Aida Amini and Saadia Gabriel and Shanchuan Lin and Rik Koncel-Kedziorski and Yejin Choi and Hannaneh Hajishirzi},
  booktitle={NAACL},
  year={2019},
}

@article{Clark2018Arc,
  title={Think you have solved question answering? try arc, the ai2 reasoning challenge},
  author={Clark, Peter and Cowhey, Isaac and Etzioni, Oren and Khot, Tushar and Sabharwal, Ashish and Schoenick, Carissa and Tafjord, Oyvind},
  journal={arXiv preprint arXiv:1803.05457},
  year={2018}
}

@article{Lin2021TruthfulQAMH,
  title={TruthfulQA: Measuring How Models Mimic Human Falsehoods},
  author={Lin, Stephanie and Hilton, Jacob and Evans, Owain},
  journal={ACL},
  year={2021}
}

@article{Cobbe2021kl,
  title={Training Verifiers to Solve Math Word Problems},
  author={Karl Cobbe and Vineet Kosaraju and Mohammad Bavarian and Mark Chen and Heewoo Jun and Lukasz Kaiser and Matthias Plappert and Jerry Tworek and Jacob Hilton and Reiichiro Nakano and Christopher Hesse and John Schulman},
  journal={NeurIPS},
  year={2021},
}

@article{Lin2023AWQAW,
  title={AWQ: Activation-aware Weight Quantization for On-Device LLM Compression and Acceleration},
  author={Lin, Ji and Tang, Jiaming and Tang, Haotian and Yang, Shang and Chen, Wei-Ming and Wang, Wei-Chen and Xiao, Guangxuan and Dang, Xingyu and Gan, Chuang and Han, Song},
  journal={MLSys},
  year={2024}
}

@article{devlin2018bert,
  title={Bert: Pre-training of deep bidirectional transformers for language understanding},
  author={Devlin, Jacob},
  journal={arXiv preprint arXiv:1810.04805},
  year={2018}
}

@article{kaplan2020scaling,
  title={Scaling laws for neural language models},
  author={Kaplan, Jared and McCandlish, Sam and Henighan, Tom and Brown, Tom B and Chess, Benjamin and Child, Rewon and Gray, Scott and Radford, Alec and Wu, Jeffrey and Amodei, Dario},
  journal={arXiv preprint arXiv:2001.08361},
  year={2020}
}

@article{touvron2023llama,
  title={Llama: Open and efficient foundation language models},
  author={Touvron, Hugo and Lavril, Thibaut and Izacard, Gautier and Martinet, Xavier and Lachaux, Marie-Anne and Lacroix, Timoth{\'e}e and Rozi{\`e}re, Baptiste and Goyal, Naman and Hambro, Eric and Azhar, Faisal and others},
  journal={arXiv preprint arXiv:2302.13971},
  year={2023}
}

@article{touvron2023llama2,
  title={Llama 2: Open foundation and fine-tuned chat models},
  author={Touvron, Hugo and Martin, Louis and Stone, Kevin and Albert, Peter and Almahairi, Amjad and Babaei, Yasmine and Bashlykov, Nikolay and Batra, Soumya and Bhargava, Prajjwal and Bhosale, Shruti and others},
  journal={arXiv preprint arXiv:2307.09288},
  year={2023}
}

@article{zhang2022opt,
  title={Opt: Open pre-trained transformer language models},
  author={Zhang, Susan and Roller, Stephen and Goyal, Naman and Artetxe, Mikel and Chen, Moya and Chen, Shuohui and Dewan, Christopher and Diab, Mona and Li, Xian and Lin, Xi Victoria and others},
  journal={arXiv preprint arXiv:2205.01068},
  year={2022}
}

@misc{vicuna2023,
    title = {Vicuna: An Open-Source Chatbot Impressing GPT-4 with 90\%* ChatGPT Quality},
    url = {https://lmsys.org/blog/2023-03-30-vicuna/},
    author = {Chiang, Wei-Lin and Li, Zhuohan and Lin, Zi and Sheng, Ying and Wu, Zhanghao and Zhang, Hao and Zheng, Lianmin and Zhuang, Siyuan and Zhuang, Yonghao and Gonzalez, Joseph E. and Stoica, Ion and Xing, Eric P.},
    month = {March},
    year = {2023}
}

@article{jiang2023mistral,
  title={Mistral 7B},
  author={Jiang, Albert Q and Sablayrolles, Alexandre and Mensch, Arthur and Bamford, Chris and Chaplot, Devendra Singh and Casas, Diego de las and Bressand, Florian and Lengyel, Gianna and Lample, Guillaume and Saulnier, Lucile and others},
  journal={arXiv preprint arXiv:2310.06825},
  year={2023}
}

@inproceedings{li2023losparse,
  title={Losparse: Structured compression of large language models based on low-rank and sparse approximation},
  author={Li, Yixiao and Yu, Yifan and Zhang, Qingru and Liang, Chen and He, Pengcheng and Chen, Weizhu and Zhao, Tuo},
  booktitle={ICML},
  year={2023},
}

@inproceedings{noach2020compressing,
  title={Compressing pre-trained language models by matrix decomposition},
  author={Noach, Matan Ben and Goldberg, Yoav},
  booktitle={AACL},
  year={2020}
}

@article{chen2023ternary,
  title={Ternary Singular Value Decomposition as a Better Parameterized Form in Linear Mapping},
  author={Chen, Boyu and Chen, Hanxuan and He, Jiao and Sun, Fengyu and Jui, Shangling},
  journal={arXiv preprint arXiv:2308.07641},
  year={2023}
}

@misc{eval-harness,
  author       = {Gao, Leo and Tow, Jonathan and Abbasi, Baber and Biderman, Stella and Black, Sid and DiPofi, Anthony and Foster, Charles and Golding, Laurence and Hsu, Jeffrey and Le Noac'h, Alain and Li, Haonan and McDonell, Kyle and Muennighoff, Niklas and Ociepa, Chris and Phang, Jason and Reynolds, Laria and Schoelkopf, Hailey and Skowron, Aviya and Sutawika, Lintang and Tang, Eric and Thite, Anish and Wang, Ben and Wang, Kevin and Zou, Andy},
  title        = {A framework for few-shot language model evaluation},
  month        = 07,
  year         = 2024,
  publisher    = {Zenodo},
  version      = {v0.4.3},
  doi          = {10.5281/zenodo.12608602},
  url          = {https://zenodo.org/records/12608602}
}

@article{taori2023alpaca,
  title={Alpaca: A strong, replicable instruction-following model},
  author={Taori, Rohan and Gulrajani, Ishaan and Zhang, Tianyi and Dubois, Yann and Li, Xuechen and Guestrin, Carlos and Liang, Percy and Hashimoto, Tatsunori B},
  journal={Stanford Center for Research on Foundation Models. https://crfm. stanford. edu/2023/03/13/alpaca. html},
  year={2023}
}

@article{marcus1993building,
  title={Building a large annotated corpus of English: The Penn Treebank},
  author={Marcus, Mitch and Santorini, Beatrice and Marcinkiewicz, Mary Ann},
  journal={Computational linguistics},
  year={1993}
}

\newpage
\appendix
\onecolumn
\section{Appendix}
\label{sec:appendix}
\subsection{Functions and Equations}
\label{sec:appendix_func}

\begin{figure}[ht]
% \vskip- 0.2in
\begin{center}
\begin{tikzpicture}
\begin{axis}[
    axis lines = center,
    xlabel = $z$,
    ylabel = {$S$},
    ymin=-0.1, ymax=2.6,
    xmin=-5, xmax=5,
    xtick={-4,-3,...,4},
    ytick={0.5,1.0,1.5,2.0},
    width=10cm,
    height=8cm,
    samples=200,
    domain=-5:5,
    legend style={at={(0.95,0.95)},anchor=north east}
]
\addplot[dotted,cyan,thick] {1/(1+exp(-x))};
\addplot[smooth,orange,very thick,domain=-2:2] {2/(1+exp(-10*x+ln(3)))};

\node at (axis cs:3.5,1.2) [font=\small,color=cyan] {$\sigma(z)=\frac{1}{1+e^{-z}}$};
\node at (axis cs:3.5,2.2) [font=\small,color=orange] {$S=\frac{2}{1+e^{-10z+ln3}}$};

\end{axis}
\end{tikzpicture}
%重要性得分$s$的函数示意图。橙色为本文使用的函数，蓝色为标准的\sigma函数。
\caption{Function graph of the importance scores $S$. The \textcolor{orange}{orange function} denotes the $S$ used in SoCo, while the \textcolor{lightblue}{blue function} represents the standard sigmoid function.}
\label{fig:score_function}
\end{center}
\vskip -0.2in
\end{figure}
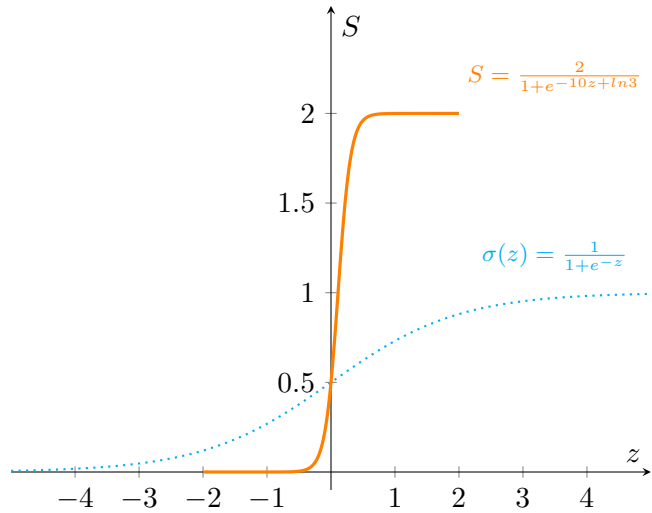

\noindent \textbf{Function graph of importance scores $S$.} As shown in Fig.~\ref{fig:score_function}~(with $\lambda_m=2$ and $\lambda_s=10$ in SoCo), the function $S$ facilitates performance compensation and rapid convergence by tuning $\lambda_m$ and $\lambda_s$. Specifically, $\lambda_m=2$ confines the overall numerical range of $S$ to $[0,2]$, allowing $S$ to compensate for the information loss incurred by pruned components with values exceeding 1.

Meanwhile, $\lambda_s$ modulates the steepness of the sigmoid-like function $S$. As shown in Eq.~\ref{eq:svd_deriv}, the derivative $\frac{dS}{dz}$ is directly proportional to $\lambda_s$, indicating that a larger $\lambda_s$ yields a faster response to changes in $z$, thereby accelerating training convergence.

\begin{equation}
\frac{dS}{dz} = \frac{\lambda_m \lambda_s (2\lambda_m-1)e^{-\lambda_s z}}{\left[1+(2\lambda_m-1)e^{-\lambda_s z}\right]^2}. 
\label{eq:svd_deriv}
\end{equation}

The term $\ln(2\lambda_m-1)$ in Eq.~\ref{eq:svd_a} is included to ensure that the function yields the same value as the sigmoid function at $z=0$. This consistency makes it easy to set a threshold for $S$.

\begin{figure}[ht]
\vskip -0.2in
\begin{center}
\begin{tikzpicture}[xscale=5,yscale=12]
\draw[->] (0,0) --(1.3,0) node[right] {$S$};
\draw[->] (0,0) --(0,0.35) node[above] {$L_{spa}$};
\foreach \x in {0,0.25,0.5,0.75,1,1.25}{\draw(\x,0)--(\x,-0.0125)node[below,outer sep=2pt,font=\tiny]at(\x,0){\x};}
\foreach \y in {0,0.05,0.10,0.15,0.20,0.25,0.30}{\draw(0,\y)--(-0.025,\y)node[left,outer sep=2pt,font=\tiny]at(0,\y){\y};}
\draw[smooth,samples=200,domain =0:1.3,line width=2pt,color=orange] plot (\x ,{(0.5-abs(\x-0.5))^2)*(\x<1)});
\draw node[font=\small,color=orange] at (0.8,0.3){$L_{spa}=\left[\left(0.5-\left|S-0.5\right|\right)\times\mathbf{1}\left(S<1\right)\right]^2$};
\draw[dotted,samples=200,domain =0:1.3,line width=1pt,color=cyan] plot (\x ,{(0.5-abs(\x-0.5))^2)});
\draw node[font=\small,color=cyan] at (1.,0.14){$L_{spa}=\left[\left(0.5-\left|S-0.5\right|\right)\right]^2$};
\end{tikzpicture}
%在$\mathcal L_{spa}$中，使得分趋于0或1的函数的示意图。橙色的函数为本文使用的函数，只在0~1范围内生效，蓝色的函数会让大于1的得分趋于1。
\caption{Illustration of the function of $spa$ in $\mathcal L_{spa}$ that drive importance scores toward 0 or 1. The \textcolor{orange}{orange function} used in SoCo converges toward 0 or 1 for scores within the range [0,1] and has no effect on scores above 1. And the \textcolor{lightblue}{blue function} ensures that scores greater than 1 are gradually pulled back toward 1.}
\label{fig:sparsity_function}
\end{center}
\vskip -0.2in
\end{figure}
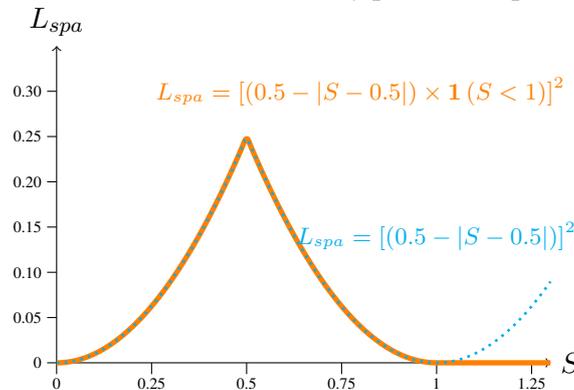

\noindent \textbf{Function graph of $spa$ in $L_{spa}$.} The function $spa$ in $L_{spa}$  drives each importance score $s$ in $S$ toward 0 when $S \leq 0.5$ and toward 1 when $0.5 < S \le 1$. Moreover, as illustrated by the orange curve in Fig.~\ref{fig:sparsity_function}, $spa^d_l$ imposes a zero gradient for $S > 1$ in Eq.~\ref{eq:l_sparsity_d}, thereby allowing $S$ to exceed 1. This mechanism enables compensation for the information loss induced by pruned components.

% \noindent \textbf{Gradient of $\mathcal L_{dec}$.} The gradient of $\mathcal L_{dec}$ with respect to $S^d_l $ is given by:
% \begin{small}
% \begin{equation}
% \label{eq:l_decreasing_d}
% \frac{\partial \mathcal L_{dec}}{\partial S^d_l} =  \frac{
%   in_l+out_l
% }{
%   \sum\limits_{l=1}^{L} \left(in_l \times out_l\right)
% }
% \end{equation}
% \end{small}
% This gradient is always positive, leading to a consistent decrease in $S$. 

Therefore, the gradient of $L_{spa}$ with respect to $S^d_l$ is defined piecewise to capture different behaviors depending on the value of $S^d_l$:

\begin{equation}
\label{eq:l_sparsity_d}
\frac{\partial L_{spa}}{\partial S^d_l} =
\begin{cases}
\frac{2 S^d_l}{LD}, & \text{if } S^d_l \leq 0.5, \\
\frac{2 (S^d_l-1)}{LD}, & \text{if } 0.5 < S^d_l \leq 1.0, \\
0, & \text{if } 1.0 < S^d_l.
\end{cases}
\end{equation}

% \begin{small}
% \begin{subnumcases}{\label{eq:l_sparsity_d} \frac{\partial L_{spa}}{\partial S^d_l} = }
%  \frac{2 S^d_l}{LD}, & $S^d_l \leq 0.5$ \label{eq:l_sparsity_d_a} \\
%  \frac{2 (S^d_l-1)}{LD}, & $0.5 < S^d_l \leq 1.0$ \label{eq:l_sparsity_d_b} \\
%  0, & $1.0 < S^d_l$ \label{eq:l_sparsity_d_c}
% \end{subnumcases}
% \end{small}

This formulation encourages sparsity by penalizing values closer to the midpoint of the interval $[0,1]$, while eliminating gradients for the values larger then 1, thereby promoting the compensation mentioned in the paper. 

% \newpage
\subsection{Algorithm implementation}
\begin{algorithm}
%\caption{移除小于0.5阈值的奇异值和相应的参数}
\caption{Removal of singular values whose importance scores $S<0.5$.}
\label{alg:ab}
\begin{lstlisting}[style=custompython]
#pytorch code
idx = (S>=0.5).nonzero().flatten()
U_k = U.index_select(1, idx)  # U_k stands for Uk in equation (2)
V_k = V.index_select(0, idx)  # V_k stands for Vk in equation (2)
Sigma_k = Sigma.index_select(0, idx)  # Sigma_k stands for Sigmak in equation (2)
S_k = S.index_select(0, idx)  # S_k stands for S', importance score after selection
\end{lstlisting}
\end{algorithm}

% \begin{algorithm}
% \vskip -0.3in
%\caption{修改重要性得分$s$梯度为1或-1 in $\mathcal L_{dec}$ and $\mathcal L_{spa}$，且奇异值越大，梯度值越小，得分降低的越慢。}
%\caption{Modify the importance score $S$ gradient to 1 or -1 in $\mathcal L_{dec}$ and $\mathcal L_{spa}$. And the larger the singular value, the smaller the gradient value, resulting in a slower decrease in importance score.}
%\caption{Modify the importance scores $S$'s gradient to 1 or -1 in $\mathcal L_{dec}$. The larger the importance score, the smaller the gradient value $1/w$ in line 26, resulting in a slower decrease in score.}
%\label{alg:BackSign}
%\begin{lstlisting}[style=custompython]
%#pytorch code for modifying gradient back-propagation of 
%class BackWeight(torch.autograd.Function):
%    @staticmethod
%    def forward(ctx,x,w):
%        ctx.save_for_backward(w)
%        return x
%    @staticmethod
%    def backward(ctx, grad_out):
%        w, = ctx.saved_tensors
%        grad = grad_out * w.to(grad_out.device)
 %       return grad,None
        
%class BackSign(torch.autograd.Function):
%    @staticmethod
%    def forward(ctx,x):
%        return x
%    @staticmethod
%    def backward(ctx, grad_out):
%%        return grad_out.sign()
        
%# S stands for the importance scores
%# indices range from len(S)-1 to 0 with 1 as interval 
%idx = torch.arange(S.numel()-1,-1,-1)
%t = 100000
%w = 1+(t-1)*idx/(S.numel()-1)
%# Take the reciprocal, make bigger S receive a smaller gradient
%w = 1/w

%S = BackSign.apply(BackWeight.apply(S,w))
%\end{lstlisting}
%\end{algorithm}
\subsection{Implementation Details}
\noindent \textbf{Training Settings.}  
The WikiText-2 dataset is concatenated and segmented into samples of 1024 tokens. The hyperparameters $\lambda_m$=2.0 and $\lambda_s$=10.0 are used as the default values in $S$. We optimize the models using AdamW with default parameters~(weight decay = 0.0, $\beta_1=0.9$, $\beta_2=0.999$, and $\epsilon=1\times10^{-8}$). Different learning rates for compressing various models are detailed in Tab.~\ref{tab:learning_rate}. A cosine scheduler with a warm-up ratio of 0.05 is applied, and the maximum gradient norm is set to 1.0. Training is performed with a batch size of 1 using BFloat16 precision and gradient checkpointing to optimize GPU memory usage. Most experiments run on a single A100 GPU; however, for LLaMA-30B, we employ FSDP~(FULL\_SHARD mode) across 4 A100 GPUs.

\noindent \textbf{Learning Rate for Different Settings.} To achieve optimal compression results, we select different learning rates based on the model and the target compression ratio. The specific values are provided in Tab.~\ref{tab:learning_rate}.

\begin{table}[ht]
%不同模型上和不同压缩率时的学习率。
\caption{Learning Rate across Different Models and Compression Ratio.}
\vspace{-3mm}
\label{tab:learning_rate}
\begin{center}
%\resizebox{0.48\textwidth}{!}{
\begin{small}
\begin{tabular}{l|ccc}
\toprule
\textsc{Model}  & Target Compression Ratio & Learning Rate  \\
\hline
LLaMA-7B   &  20\%         & 0.002  \\
LLaMA-7B   &  40\%         & 0.002  \\
LLaMA-7B   &  60\%         & 0.007  \\
LLaMA-7B   &  80\%         & 0.008  \\
LLaMA-13B   &  20\%         & 0.002  \\
LLaMA-30B   &  20\%         & 0.003  \\
LLaMA2-7B   &  20\%         & 0.002  \\
OPT-6.7B   &  20\%         & 0.002  \\
Mistral-7B   &  20\%         & 0.001  \\
Vicuna-7B   &  20\%         & 0.002  \\
\bottomrule
\end{tabular}
%}
\end{small}
\end{center}
\end{table}

\subsection{Supplementary Ablation Study}

\noindent \textbf{Deviation Term $d$.} To compensate for any deviation introduced by modifying the weight matrix $W'$, we add a trainable deviation term $d$ after the linear transformation of $W'$ in SoCo. This deviation term quantifies the average deviation and helps mitigate performance loss. Its effectiveness is demonstrated in Tab.~\ref{tab:alba_delta}.

\begin{table}[ht]
\caption{Ablation experiments of SoCo on the use of the deviation term $d$, conducted on the LLaMA-7B model at the compression ratios of 20\%, 40\%, 60\%, and 80\%.}
\vspace{-3mm}
\label{tab:alba_delta}
\begin{center}
\resizebox{0.48\textwidth}{!}{
\begin{tabular}{cl|ccc}
\toprule
\textsc{Ratio} & \textsc{Method} & WikiText-2$\downarrow$ & C4$\downarrow$   & Average$\uparrow$  \\
\hline
20\% & w/o Deviation & 6.77           & 9.92           & 52.36\%          \\
     & \textbf{w Deviation} & \textbf{6.67}  & \textbf{9.89}  & \textbf{52.48\%} \\
\hline
40\% & w/o Deviation & 9.69           & 15.56          & 45.52\%          \\
     & \textbf{w Deviation} & \textbf{8.60}  & \textbf{13.71} & \textbf{47.16\%} \\
\hline
60\% & w/o Deviation & 16.27          & 31.80          & 38.01\%          \\
     & \textbf{w Deviation} & \textbf{12.20} & \textbf{24.53} & \textbf{40.22\%} \\
\hline
80\% & w/o Deviation & 33.72          & 81.57          & 33.22\%          \\
     & \textbf{w Deviation} & \textbf{21.03} & \textbf{50.86} & \textbf{34.63\%} \\
\bottomrule
\end{tabular}
}
\end{center}
\end{table}
\noindent \textbf{Hyper Parameters $\lambda_m$ and $\lambda_s$ in $S$.} \label{hyper}
Excluding one outlier at $(\lambda_m, \lambda_s) = (2.0, 1.0)$ in Tab.~\ref{tab:ablation_lambda}, the remaining results vary only slightly, demonstrating the robustness of SoCo with respect to these hyperparameters. We adopt a relatively optimal configuration $(\lambda_m, \lambda_s) = (2.0, 10.0)$ for other experiments.

\begin{table}[ht]
\caption{Ablation study on the hyperparameters \(\lambda_m\) and \(\lambda_s\) in \(S\). Performance is evaluated via PPL on WikiText-2 and C4, and the averaged accuracy of 6 classification tasks in LM-Evaluation-Harness.
}
\vspace{-3mm}
\begin{center}
\resizebox{0.36\textwidth}{!}{
% \rowcolors{1}{}{lightgray}
\begin{tabular}{ccccc}
\toprule
\textsc{$\lambda_m$} & \textsc{$\lambda_s$} & WikiText-2$\downarrow$ & C4$\downarrow$   & Average$\uparrow$  \\
\hline
1.0          & ~~1.0                & 7.47          & 12.17          & 49.38\%          \\
1.0          & 10.0                 & 6.80          & 9.99          & 52.42\%          \\
\textbf{2.0} & \textbf{10.0} & \textbf{6.67} & \textbf{9.89} & \textbf{52.48\%} \\
4.0          & 10.0                 & 6.75          & 9.98          & 51.97\%          \\
8.0          & 10.0                 & 6.77          & 10.00         & 51.73\%          \\
2.0          & ~~1.0                & 10.30         & 17.14         & 42.52\%          \\
2.0          & ~~2.0                & 8.46          & 12.92         & 49.34\%          \\
2.0          & ~~5.0                & 6.98          & 10.27         & 51.52\%          \\
2.0          & 15.0                 & 6.73          & 9.97          & 52.34\%          \\
2.0          & 20.0                 & 6.77          & 9.99          & 52.37\%          \\
\bottomrule
\end{tabular}
}
\end{center}
\label{tab:ablation_lambda}
\end{table}

\subsection{Compression Result Observations}

\begin{figure*}[ht]
\vspace{-3mm}
\begin{center}
    % \begin{tabular}{@{\extracolsep{\fill}}c@{}c@{}{\extracolsep{\fill}}}
    \begin{tabular}{@{\extracolsep{\fill}}c@{\extracolsep{\fill}}c@{\extracolsep{\fill}}}
        \includegraphics[width=0.5\linewidth]{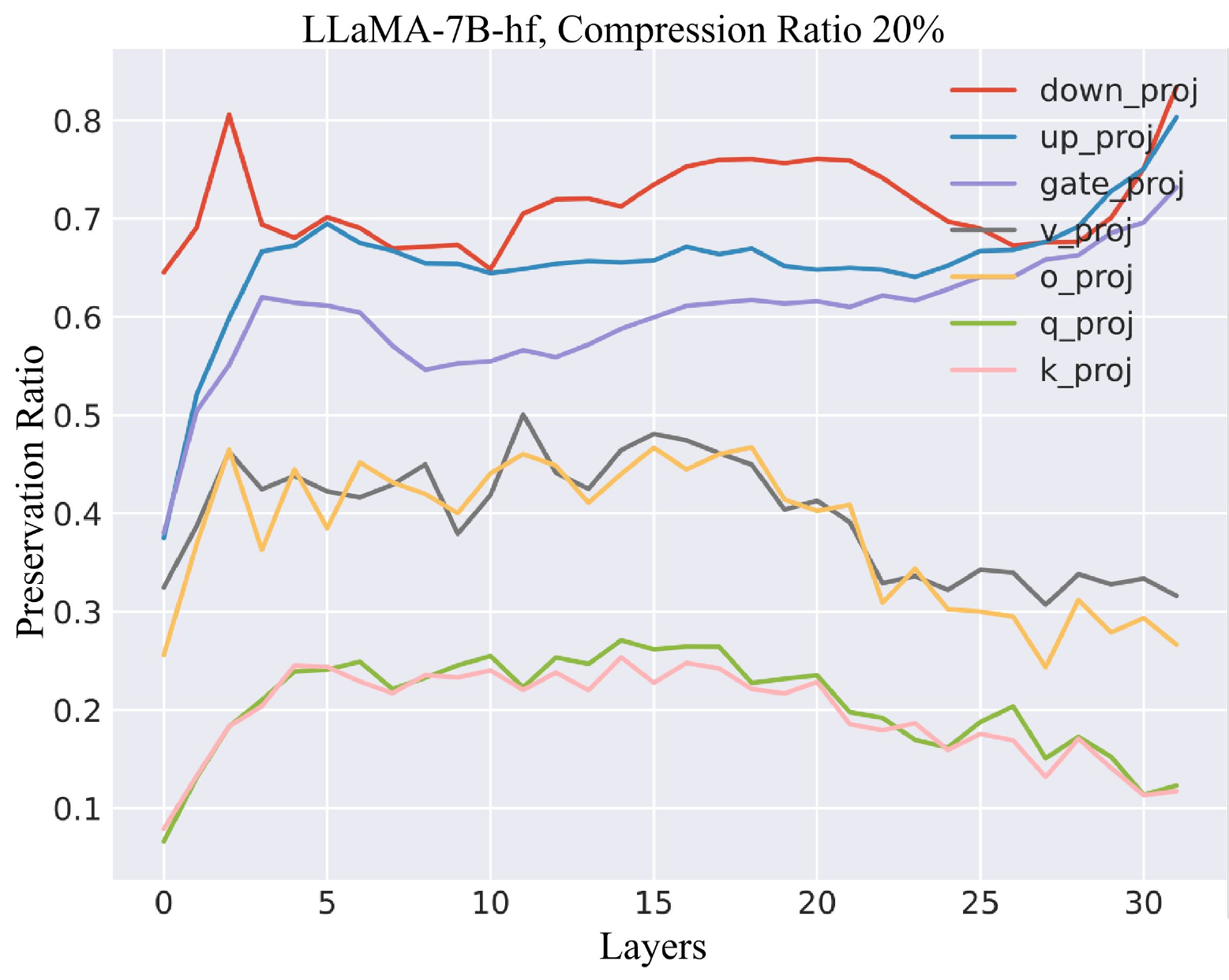} &
        \includegraphics[width=0.5\linewidth]{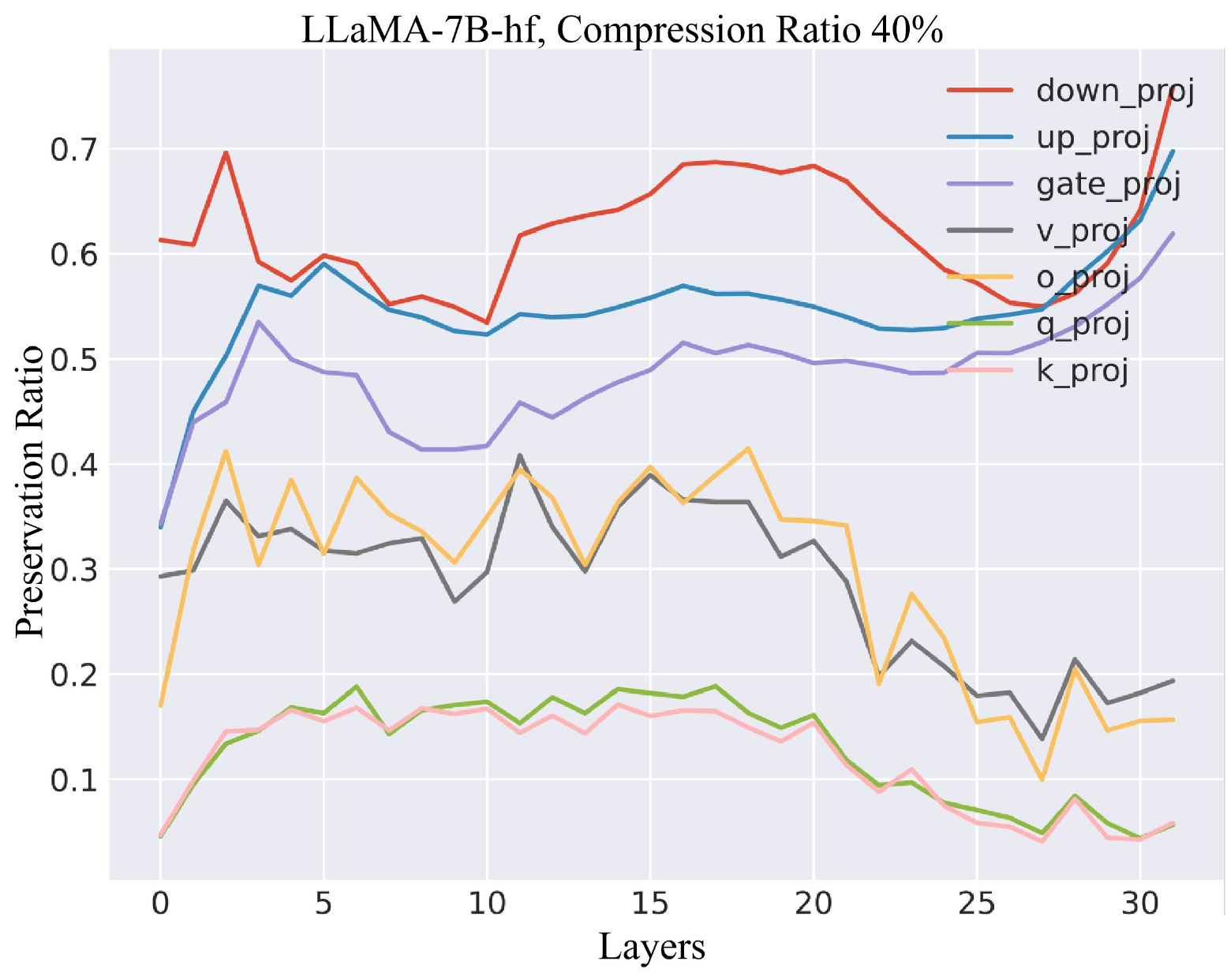} \\
        \includegraphics[width=0.5\linewidth]{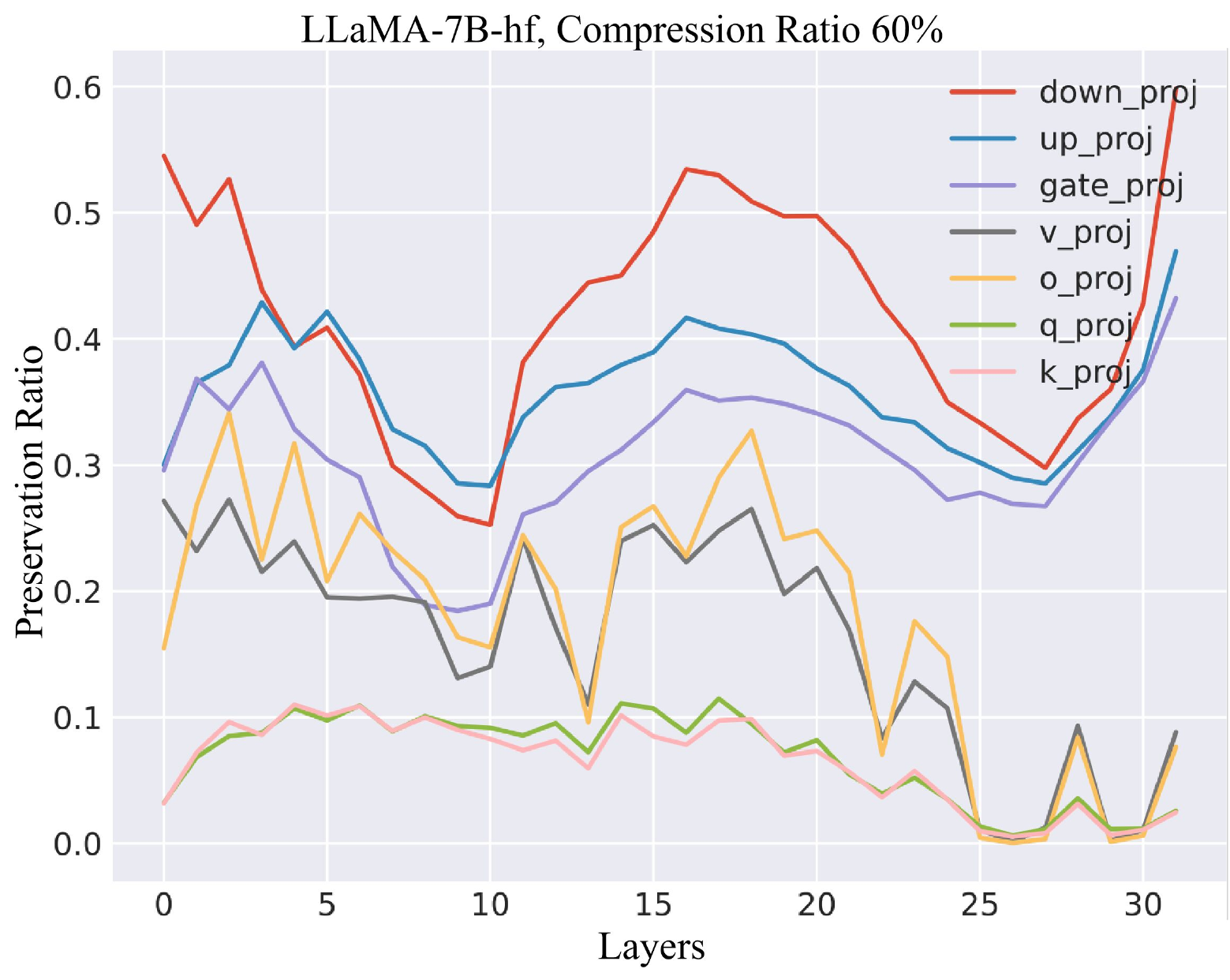} & 
        \includegraphics[width=0.5\linewidth]{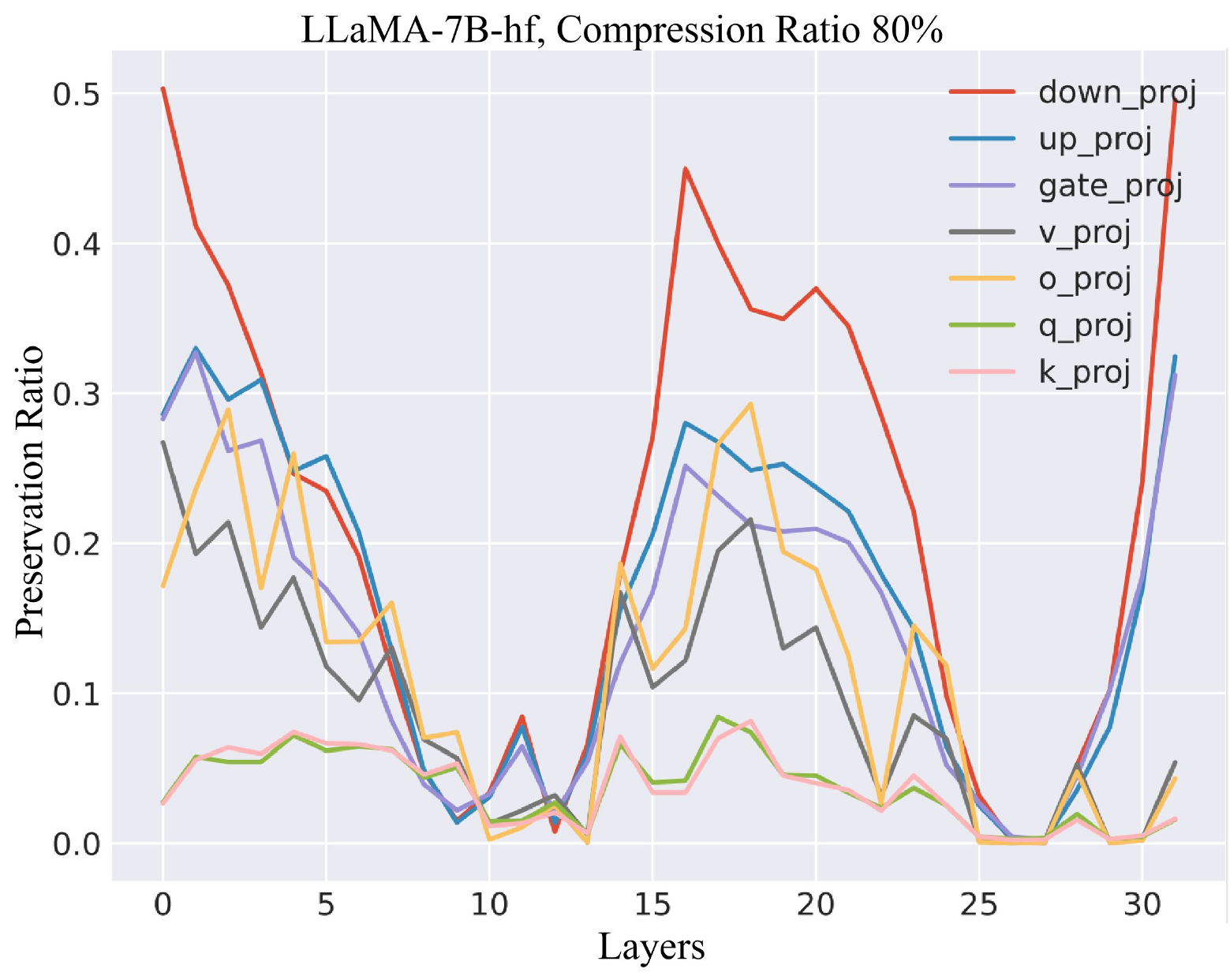} \\
        
    \end{tabular}
%在不同家族系列的模型上，在不同压缩率时，经过压缩处理后，每种linear module和每层layer的剩余组件所占的比例。
\caption{The four subplots depict the preservation ratio of decomposed components in SoCo for LLaMA-7B at the compression ratios of 20\%, 40\%, 60\%, and 80\%. 
% In each subplot, every horizontal axis point corresponds to a transformer layer, and the seven colored lines represent the remnant proportions of the decomposed SVD components for each distinct linear module’s weight matrix within that layer after compression.
}
\label{fig:remnant_ratio}
\end{center}
\vskip -0.2in
\end{figure*}

\noindent \textbf{Decomposed Components Preservation Ratio Across All Layers in LLaMA at Different Compression Ratios.} 
The four subplots in Fig.~\ref{fig:remnant_ratio} show the proportion of preserved components across the layers of LLaMA-7B at compression ratios of 20\%, 40\%, 60\%, and 80\%. In each subplot, each point on the horizontal axis represents a transformer layer, and the plot shows seven colored lines—each corresponding to a different linear module’s weight matrix within that layer. The vertical values of these lines indicate the preservation proportion of decomposed components from the original SVD for each module.

Overall, the preservation ratio of decomposed components for the FFN linear modules~(down\_pro, up\_proj, gate\_proj) is higher than that for the Multi-Head Attention linear modules~(q\_proj, k\_proj, v\_proj, o\_proj). Moreover, as the compression ratio increases, the decrease in the preservation ratio is less pronounced for the Multi-Head Attention modules than for the FFN modules. 

\noindent \textbf{Decomposed Components Preservation Ratio Across All Layers in Different LLMs at 20\% Compression Ratio.} 

\begin{figure*}[ht]
\vskip 0.2in
\begin{center}
    % \begin{tabular}{@{\extracolsep{\fill}}c@{}c@{}c@{}c@{\extracolsep{\fill}}l}
    \begin{tabular}{@{\extracolsep{\fill}}c@{\extracolsep{\fill}}c@{\extracolsep{\fill}}}
        \includegraphics[width=0.5\linewidth]{llama-7b-hf_c0.8_e2lr0.002_wA.pdf} &
        \includegraphics[width=0.5\linewidth]{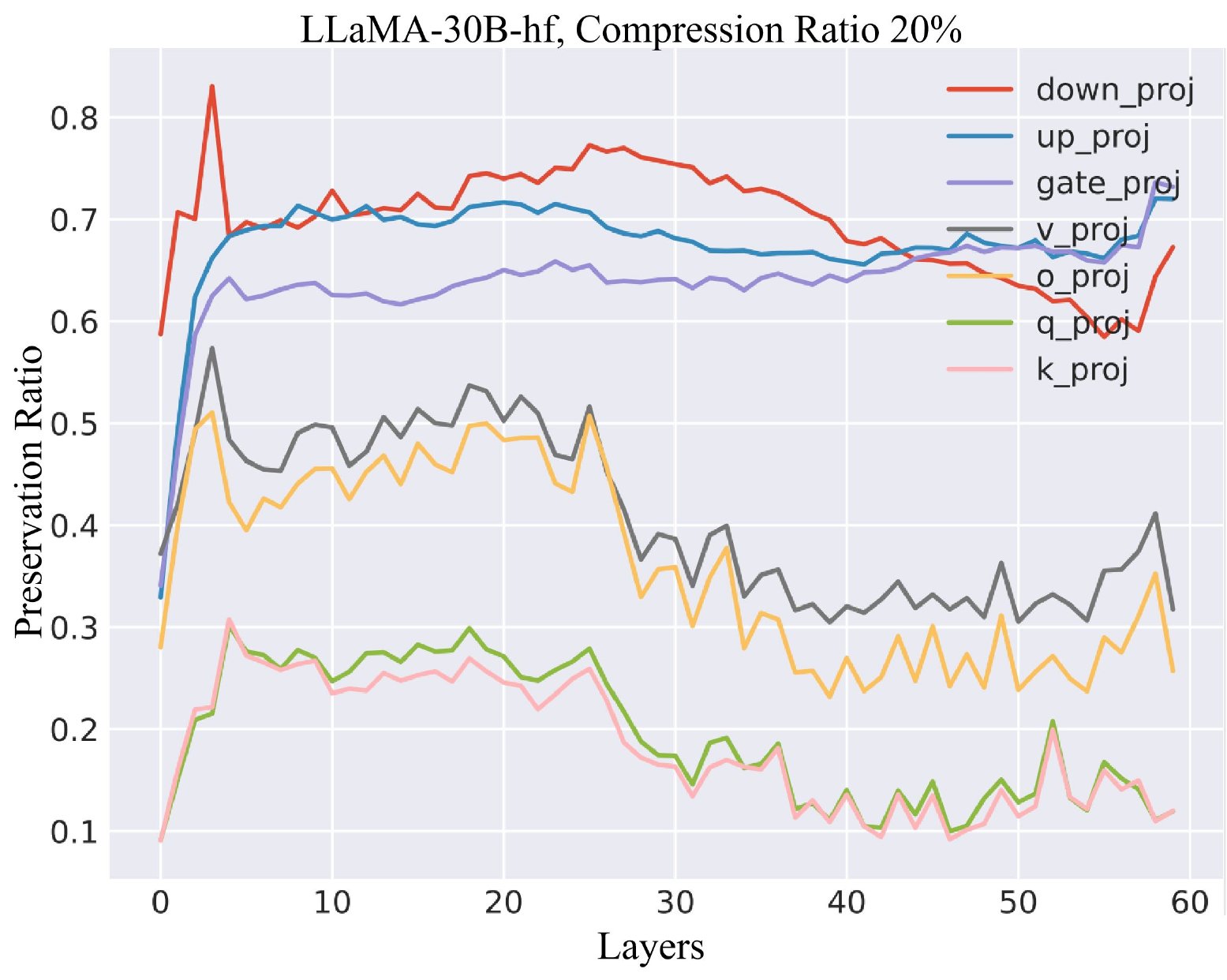} \\
        \includegraphics[width=0.5\linewidth]{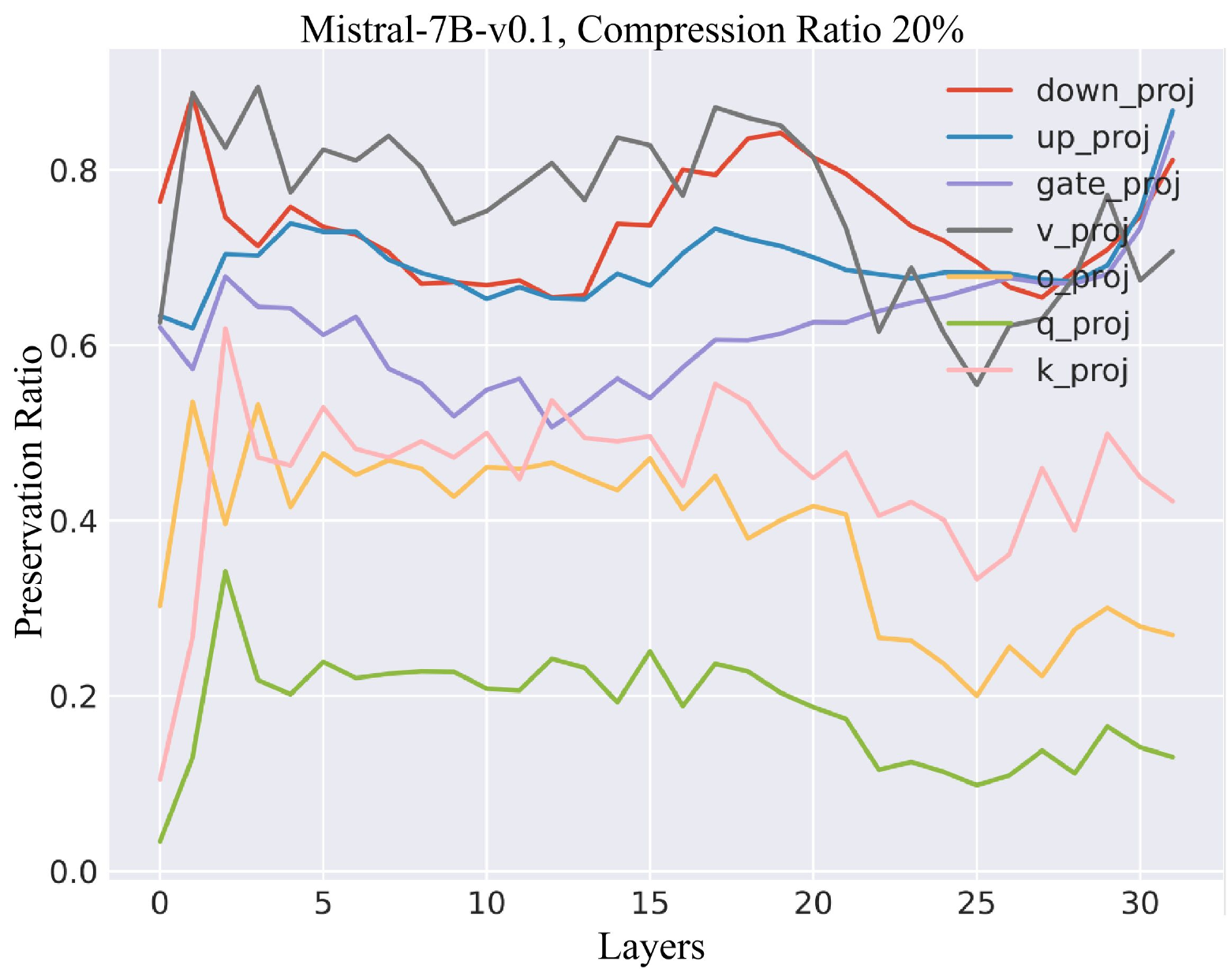} & 
        \includegraphics[width=0.5\linewidth]{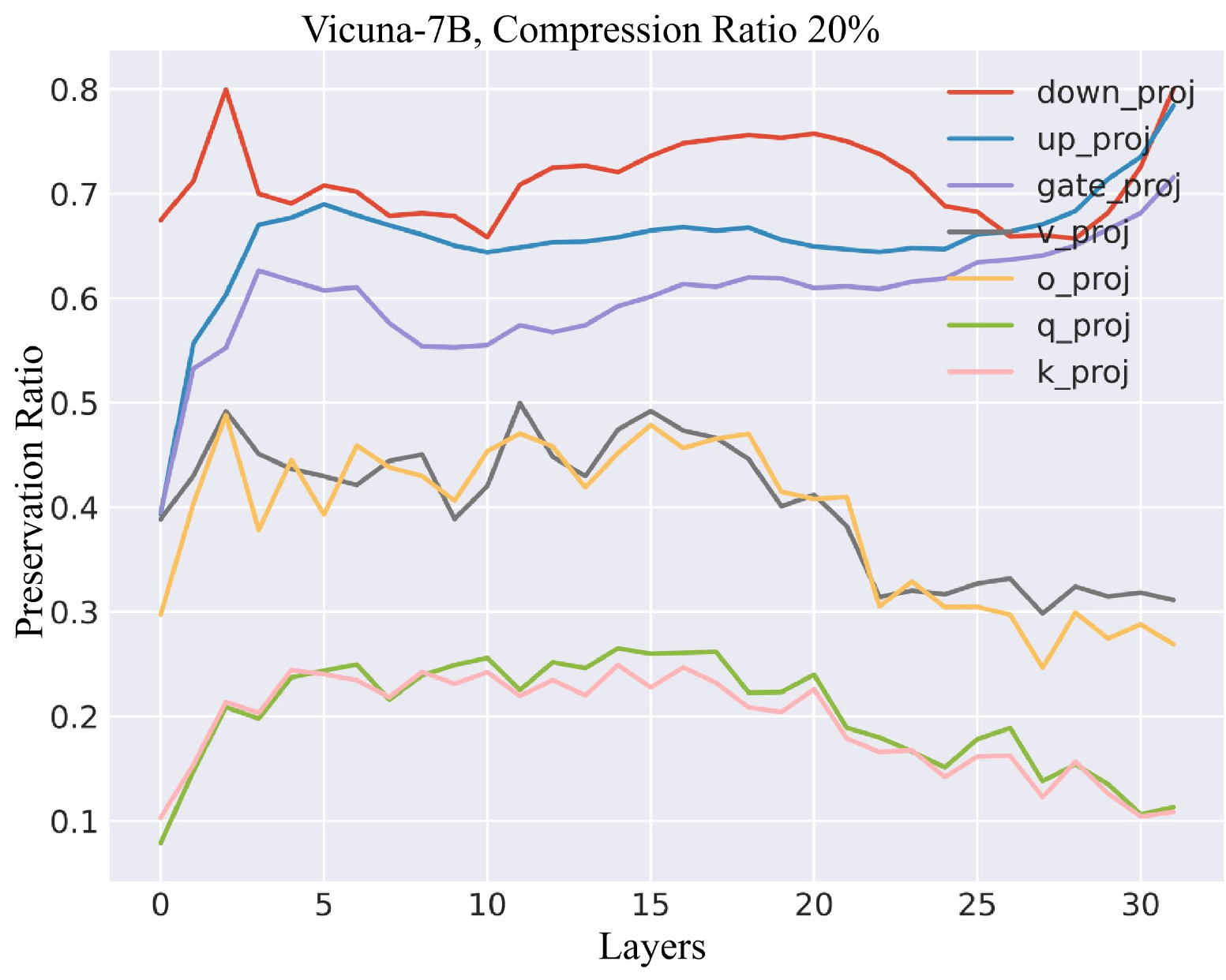} \\
        \includegraphics[width=0.5\linewidth]{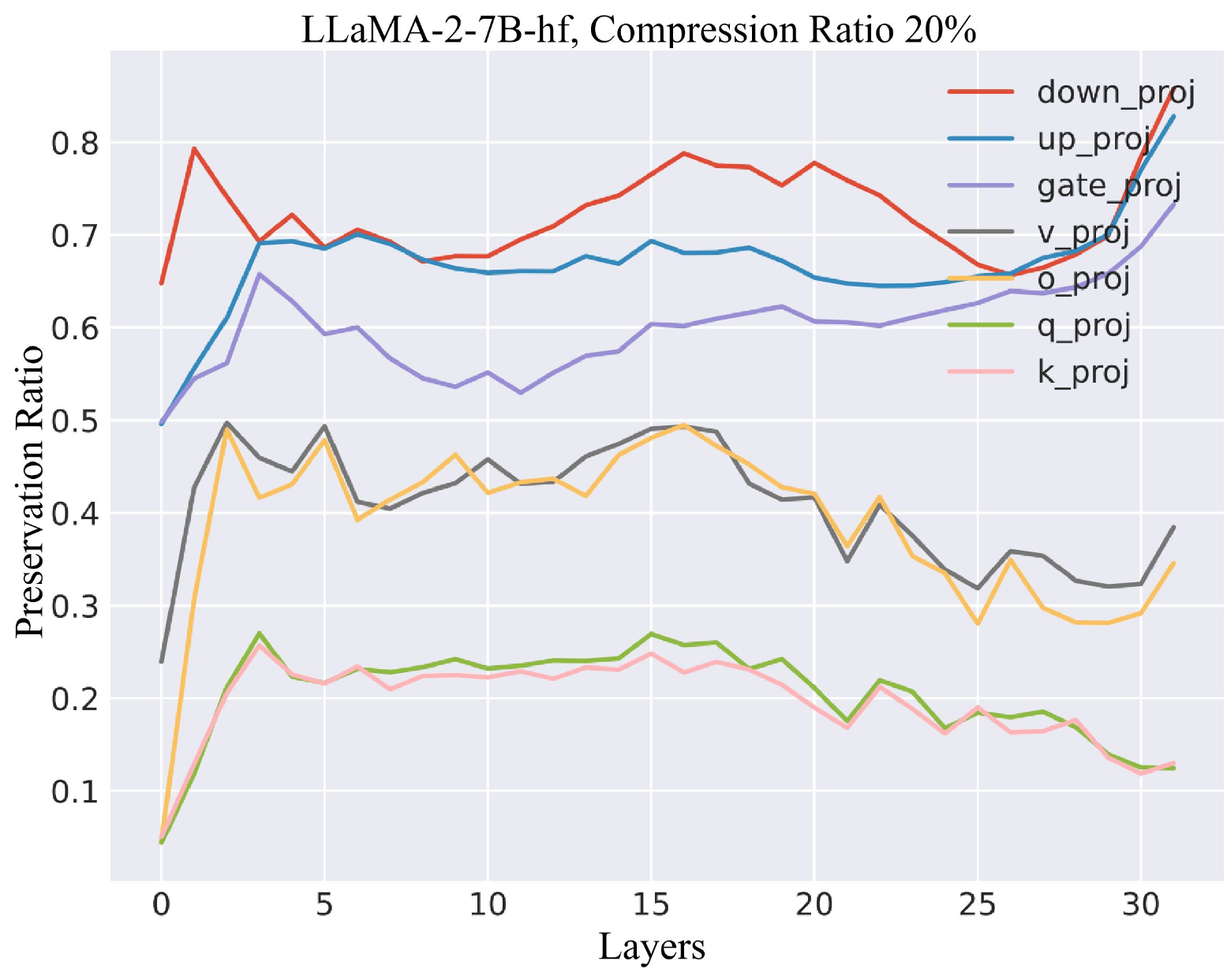} &
        \includegraphics[width=0.5\linewidth]{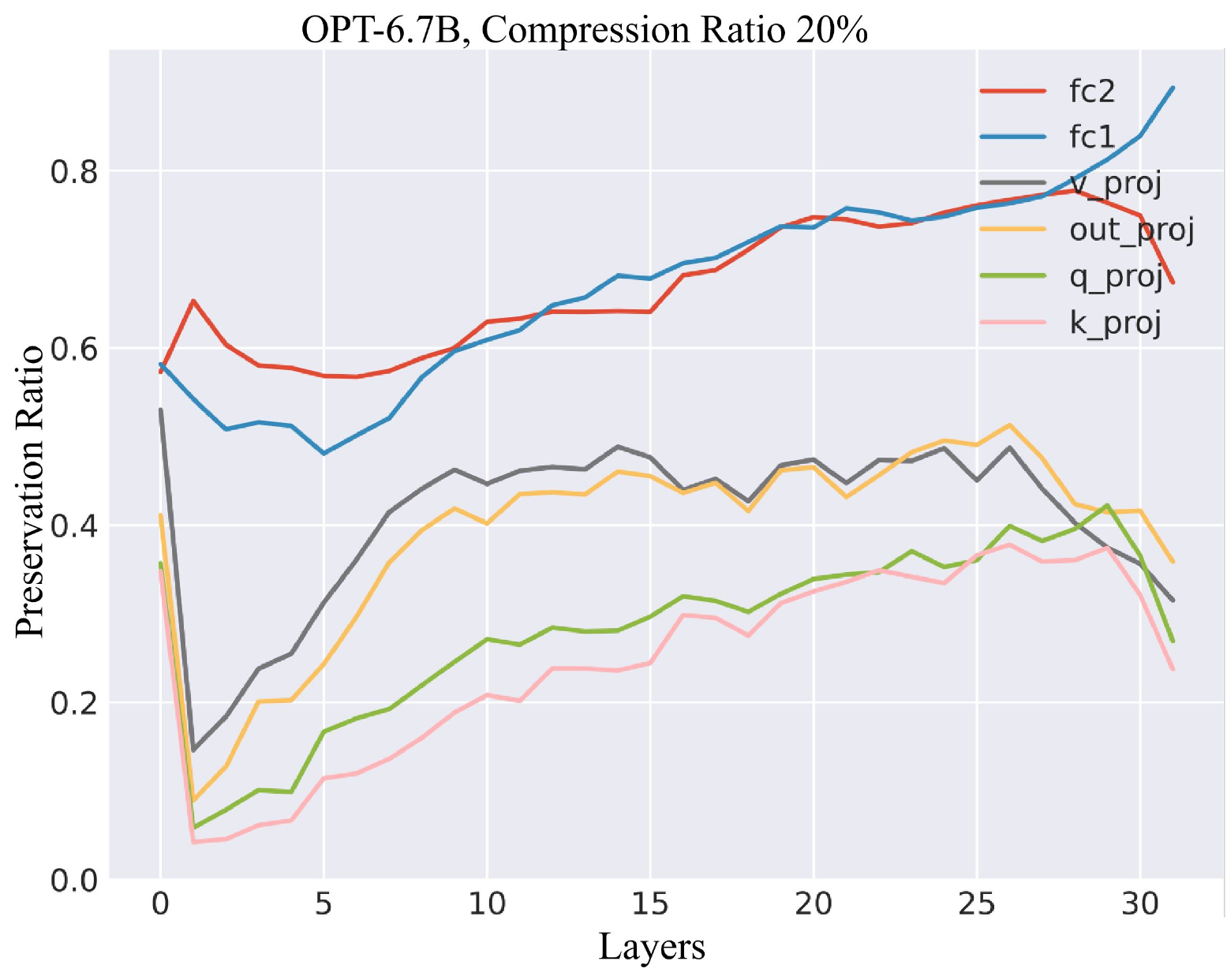} \\
        
    \end{tabular}
%在不同家族系列的模型上，在不同压缩率时，经过压缩处理后，每种linear module和每层layer的剩余组件所占的比例。
\caption{The six subplots depict the preservation ratio of decomposed components in SoCo across all transformer layers and linear modules in various LLMs at 20\% compression ratio.}
\label{fig:remnant_model}
\end{center}
\vskip -0.2in
\end{figure*}

Fig.~\ref{fig:remnant_model} presents the preserved component proportions from SoCo across the transformer layers of various models—LLaMA-7B, LLaMA-30B, Mistral-7B, Vicuna-7B, Llama-2-7b, and OPT-6.7B—all evaluated at a 20\% compression ratio. In general, the preservation ratio of decomposed components for the FFN linear modules is higher than that for the Multi-Head Attention linear modules (q\_proj, k\_proj, v\_proj, o\_proj). Furthermore, no clear pattern emerges across the various models.

\end{document}